\documentclass[letterpaper]{article} 
\usepackage[preprint]{aaai2027}
\usepackage[hyphens]{url}  
\usepackage{graphicx} 
\usepackage{natbib}  
\usepackage{caption} 
\usepackage{algorithm}
\usepackage{algorithmic}

\usepackage{newfloat}
\usepackage{listings}
\DeclareCaptionStyle{ruled}{labelfont=normalfont,labelsep=colon,strut=off} 
\floatstyle{ruled}
\newfloat{listing}{tb}{lst}{}
\floatname{listing}{Listing}
\usepackage{multirow}
\usepackage{xcolor}
\newcommand{\red}[1]{\textcolor{red}{#1}}
\usepackage{amsmath}
\usepackage{amsfonts}
\usepackage{colortbl}
\definecolor{SOTARow}{RGB}{232,240,248}
\usepackage{pifont}
\newcommand{\cmark}{\textcolor{green!60!black}{\ding{51}}}
\newcommand{\xmark}{\textcolor{red}{\ding{55}}}

\usepackage{subcaption}
\usepackage{booktabs}
\title{Echo-Forcing: Lifecycle-Aware KV Memory Management for Interactive Autoregressive Long-Video Generation}
\author{%
    Mingqiang Wu\textsuperscript{\normalfont 1,2,\textdagger}\quad
    Weilun Feng\textsuperscript{\normalfont 1,2,\textdagger}\quad
    Zhefeng Zhang\textsuperscript{\normalfont 3}\quad
    Haotong Qin\textsuperscript{\normalfont 4}\\
    Guoxin Fan\textsuperscript{\normalfont 1,2}\quad
    Xiaokun Liu\textsuperscript{\normalfont 1,2}\quad
    Zhulin An\textsuperscript{\normalfont 1}\\
    Libo Huang\textsuperscript{\normalfont 1}\quad
    Yongjun Xu\textsuperscript{\normalfont 1,6}\quad
    Chuanguang Yang\textsuperscript{\normalfont 1,*}%
}
\affiliations{%
    \textsuperscript{1}State Key Laboratory of AI Safety, Institute of Computing Technology,\\
    Chinese Academy of Sciences, Beijing, China\\
    \textsuperscript{2}University of Chinese Academy of Sciences, Beijing, China\\
    \textsuperscript{3}China University of Mining \& Technology, Beijing\\
    \textsuperscript{4}ETH Z\"urich\\
    \textsuperscript{6}Xiamen Institute of Data Intelligence, Xiamen, China%
}

\newcommand{\printauthornotes}{%
    \begingroup
    \renewcommand{\thefootnote}{\textdagger}%
    \footnotetext{Equal contribution.}%
    \renewcommand{\thefootnote}{*}%
    \footnotetext{\raggedright Corresponding author:\\
        Chuanguang Yang \mbox{\textless{}yangchuanguang@ict.ac.cn\textgreater{}}.}%
    \endgroup
}

\begin{document}

\AddToHookNext{cmd/@maketitle/after}{%
    \noindent
    \begin{minipage}{\textwidth}
        \centering
        \includegraphics[width=\linewidth]{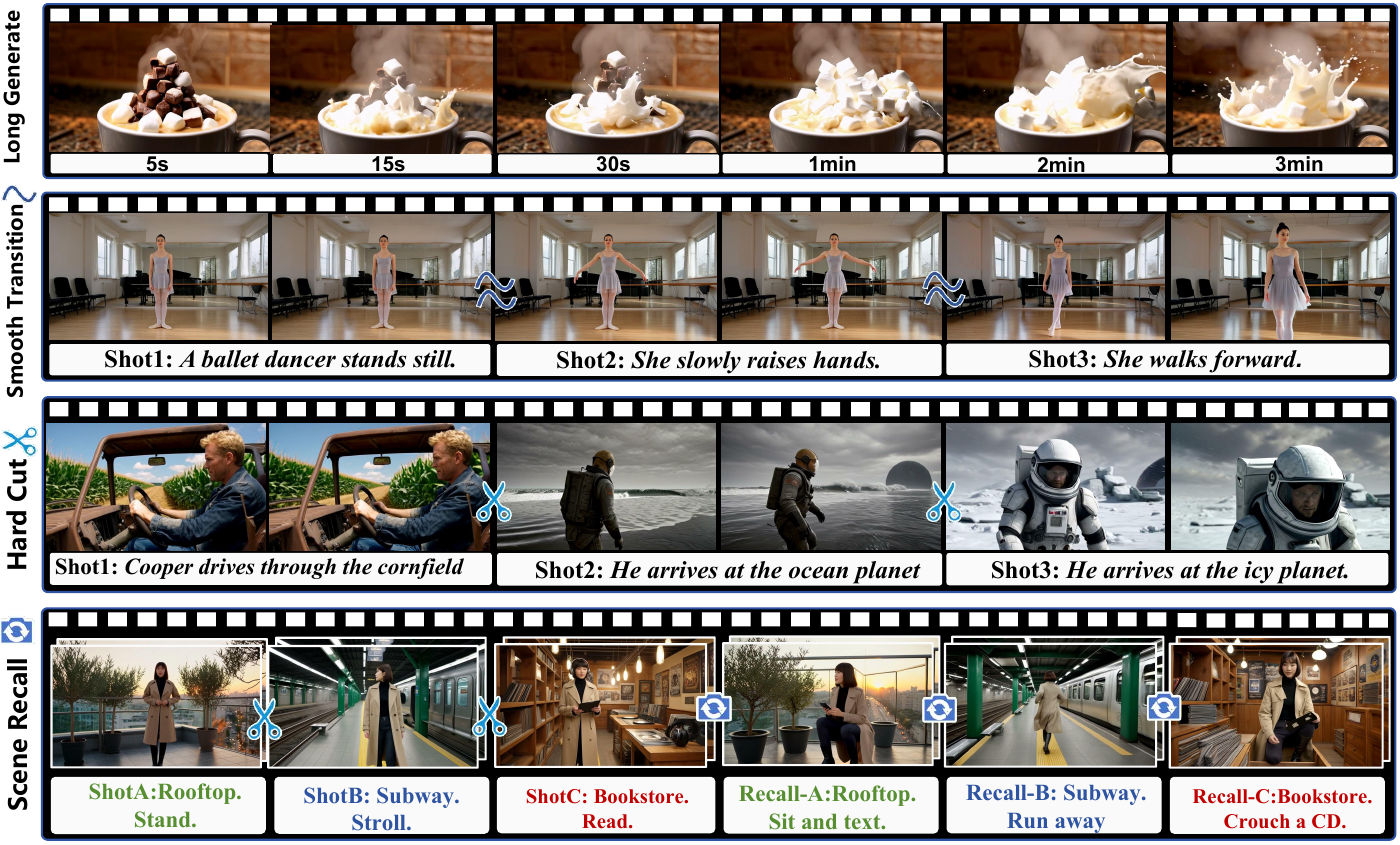}
        \captionof{figure}{\textbf{Echo-Forcing} enables autoregressive video diffusion models to support four interactive long-video generation modes: long-horizon generation, smooth transition, hard cut, and long-range scene recall, while maintaining temporal coherence and scene consistency.}
        \label{mt-t0}
    \end{minipage}\par
}
\maketitle
\printauthornotes

\begin{abstract}
Autoregressive video diffusion models enable open-ended generation through local attention and KV caching. However, existing training-free long-video methods primarily target stable extension under a single prompt, leaving prompt switching, historical scene recall, and old-scene forgetting largely unaddressed. We argue that the key challenge is not video length alone, but lifecycle-aware KV memory management: a flat cache cannot distinguish memories that should be updated during ongoing generation, retrieved at scene transitions, or decayed afterwards, causing stale-scene contamination, delayed responses to new prompts, and loss of long-range scene priors. To address this issue, we propose \textbf{Echo-Forcing}, a training-free framework that models interactive autoregressive video generation as KV memory evolution. It contains three coordinated mechanisms: (1) \textit{Hierarchical Temporal Memory}, which organizes and updates stable anchors, compressed history, and recent windows under relative RoPE; (2) \textit{Scene Recall Frames}, which construct compact spatially structured KV memories from historical scenes for transition-time retrieval; and (3) \textit{Difference-aware Memory Decay}, which selectively attenuates conflicting old KV tokens according to old--new scene discrepancies. Together, these modules implement an update--retrieve--decay lifecycle under a bounded cache budget, supporting stable long-video generation, smooth transitions, hard cuts, and long-range scene recall. Extensive evaluations on VBench-Long and interactive generation settings demonstrate that Echo-Forcing achieves the strongest overall performance in both long-video quality and interactive controllability.Our code is released in 
{\texttt{https://github.com/mingqiangWu/Echo-Forcing}}.
\end{abstract}

\begin{figure*}[t]
    \centering
    \includegraphics[width=\textwidth]{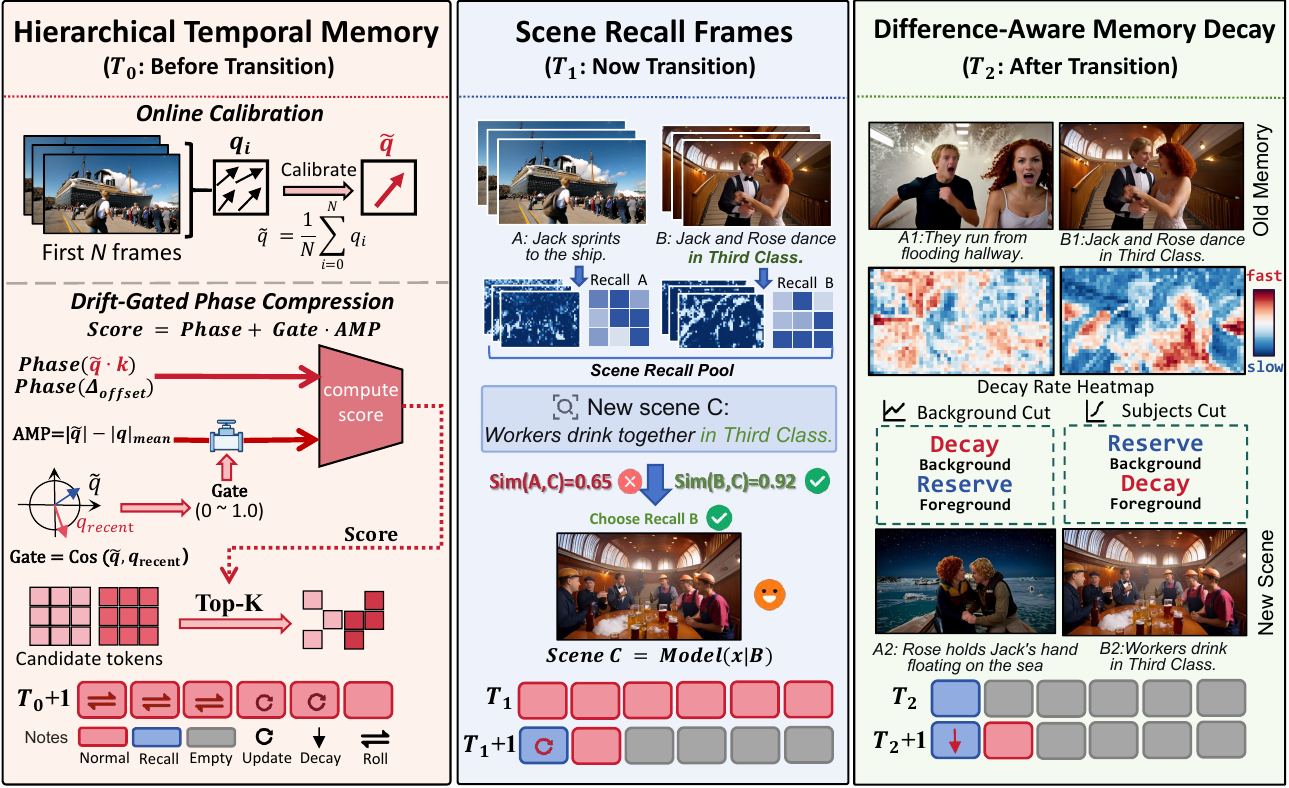}
    \caption{
    Lifecycle-aware KV memory management in Echo-Forcing.
    Hierarchical Temporal Memory operates during ongoing generation, Scene Recall Frames are activated at scene transitions, and Difference-aware Memory Decay is applied after transitions. Together, they implement an update--retrieve--decay memory lifecycle.
    }
    \label{fig:overview}
\end{figure*}

\section{Introduction}

Video generation~\citep{kong2024hunyuanvideo,wan2025wan,liu2024sora,kondratyuk2023videopoet,bar2024lumiere,guo2023animatediff,ho2022imagen,ho2022video,qvdit,feng2025quantsparse,feng2025s,feng2026s} is moving from offline short-clip synthesis toward open-ended, interactive creation. Autoregressive video diffusion models~\citep{yin2025causvid,huang2025self,cui2025selfplus,MAGI,liu2025rolling} provide a scalable paradigm for this shift: by generating videos block by block and reusing historical key--value (KV) states~\citep{zhang2023h2o,li2024snapkv,liu2023scissorhands}, they enable continuous video extension with bounded inference cost. However, error accumulation over long-horizon rollout and timely adaptation to evolving streaming instructions both hinge on managing the evolving KV cache under a limited memory budget.

Existing KV-cache methods are temporally limited, typically optimizing only one stage of the interactive generation lifecycle. Training-free long-video methods improve cache retention and compression during continuous single-prompt generation through positional adaptation~\citep{yesiltepe2025infinity}, sink-token retention~\citep{huang2025self,yi2025deep,li2026rolling}, or historical-cache compression~\citep{yi2025deep,kim2026memrope,lv2026light}. In contrast, interactive and multi-shot methods primarily intervene at the instant of prompt switching through cache re-injection~\citep{yang2025longlive}, cache flushing~\citep{yesiltepe2025infinity}, or local transition control~\citep{luo2026shotstream}.

Such flat and task-agnostic management creates unavoidable trade-offs. Retaining history stabilizes continuous generation, but can propagate stale scene semantics after a scene switch; conversely, flushing or overwriting the cache improves immediate responsiveness to new prompts, but discards historical priors needed for later scene recall. These uniform policies cannot jointly provide long-horizon stability, responsiveness, and scene recall because they ignore stage- and task-dependent memory evolution.

We therefore formulate interactive autoregressive video generation as \emph{lifecycle-aware KV memory evolution}: historical KV states are updated during ongoing generation, retrieved at scene transitions, and retained or decayed afterwards. Based on this principle, we propose \textbf{Echo-Forcing}, a training-free KV memory management framework for interactive autoregressive video diffusion. \textbf{Hierarchical Temporal Memory} maintains stable anchors, compressed history, and recent dynamics; \textbf{Scene Recall Frames} encode compact spatially structured scene memories for retrieval; and \textbf{Difference-aware Memory Decay} attenuates conflicting information according to old--new scene differences. Together, they implement an update--retrieve--decay KV memory lifecycle for interactive long-video generation.

Our contributions are summarized as follows:
\begin{itemize}
    \item \textbf{Lifecycle Formulation.}
    We formulate interactive autoregressive long-video generation as a lifecycle-aware KV memory evolution problem, unifying long-horizon stability, interactive switching, and scene recall.
    
    \item \textbf{Unified Framework.}
    We propose Echo-Forcing, a training-free KV memory management framework that coordinates KV updating, retrieval, and decay across ongoing generation, scene transitions, and post generation.
    
    \item \textbf{Stage-Coupled Modules.}
    We instantiate this framework with Hierarchical Temporal Memory, Scene Recall Frames, and Difference-aware Memory Decay, which jointly support stable maintenance, historical recall, and conflict suppression.
    
    \item \textbf{Empirical Validation.}
    Extensive experiments on long-video and interactive generation demonstrate improved long-horizon quality, scene consistency, prompt responsiveness, and scene-recall capability.
\end{itemize}

\section{Related Work}

\subsubsection{Autoregressive Video Generation.}
In recent years, high-fidelity video generation~\citep{kong2024hunyuanvideo,liu2024sora,yang2024cogvideox,wan2025wan,team2025kling,gupta2024photorealistic,feng2026worldcache} has been largely driven by bidirectional-attention DiT architectures~\citep{peebles2023scalable,ma2024latte,bao2023all}, whose denoising process jointly models the full temporal context and incurs substantial computational overhead. To enable streaming inference, CausVid~\citep{yin2025causvid} and Self-Forcing~\citep{huang2025self} distill bidirectional DiTs into causal generators~\citep{yin2024dmd}, but still suffer from degradation under length extrapolation. Subsequent training-based works extend generation horizons and interactive capabilities through long-rollout training~\citep{cui2025selfplus,liu2025rolling,yang2025longlive}, block-wise prediction~\citep{yang2025longlive}, reward distillation~\citep{lu2025reward}, and semantic--dynamic decoupling~\citep{chen2026grounded}. Recent training-free optimizations focus on positional encoding adaptation~\citep{zhao2025riflex,yesiltepe2025infinity,kim2026memrope}, KV-cache management~\citep{li2026rolling,yi2025deep,xiao2024efficient,liu2025reattention}, and attention efficiency~\citep{guo2026dummy,lv2026light}. While these methods improve long-video stability or efficiency, most still focus on continuous rollout under a single prompt.

\subsubsection{Multi-shot and Interactive Video Generation.}
Existing multi-shot video generation methods mainly target cross-shot consistency through fixed-window attention~\citep{qi2025mask,kara2025shotadapter,guo2025lct}, key-frame conditioning~\citep{zhou2024storydiffusion,xiao2025captain,he2025cut2next}, and adaptive historical memory~\citep{an2025onestory,luo2026shotstream}. Fixed-window methods tend to lose earlier shots as the window slides, whereas key-frame-based methods often rely on multi-stage generation pipelines. SwitchCraft~\citep{xu2026switchcraft} further explores interactive generation with a bidirectional video model. In contrast, autoregressive generators~\citep{yin2025causvid,huang2025self,liu2025rolling} naturally support streaming interaction by reusing historical KV caches, through which prompt updates and long-range dependencies are mediated. Existing streaming interactive methods mainly reinject new prompt semantics~\citep{yang2025longlive} or modify KV retention and RoPE temporal coordinates~\citep{yesiltepe2025infinity,chen2026grounded,shin2025motionstream,samuel2026fast}. However, these methods do not explicitly distinguish contextual transition types, limiting their adaptability to large semantic shifts and long-range memory dependencies.

\section{Methods}

\begin{table*}[t]
\centering
{\small
\setlength{\tabcolsep}{2.6pt}
\renewcommand{\arraystretch}{0.95}
\begin{tabular}{lccccccc}
\toprule
Method
& \begin{tabular}{@{}c@{}}Throughput\\FPS${\mathbf{\red{\uparrow}}}$\end{tabular}
& \begin{tabular}{@{}c@{}}Aesthetic\\ Quality${\mathbf{\red{\uparrow}}}$\end{tabular}
& \begin{tabular}{@{}c@{}}Background\\ Consistency${\mathbf{\red{\uparrow}}}$\end{tabular}
& \begin{tabular}{@{}c@{}}Imaging\\ Quality${\mathbf{\red{\uparrow}}}$\end{tabular}
& \begin{tabular}{@{}c@{}}Motion\\ Smooth.${\mathbf{\red{\uparrow}}}$\end{tabular}
& \begin{tabular}{@{}c@{}}Temporal\\ Flickering${\mathbf{\red{\uparrow}}}$\end{tabular}
& \begin{tabular}{@{}c@{}}Overall\\ Avg.${\mathbf{\red{\uparrow}}}$\end{tabular} \\
\midrule

\multicolumn{8}{c}{\textbf{Results on 60s}} \\
\midrule
Self-Forcing~\citep{huang2025self}
& \textbf{17.01} & 56.32 & \underline{96.35} & 68.26
& 98.57 & 97.94 & 83.49 \\
$\infty$-RoPE~\citep{yesiltepe2025infinity}
& \textbf{17.01} & 58.49 & 94.66 & 70.36
& 97.95 & 96.10 & 83.51 \\
Deep-Forcing~\citep{yi2025deep}
& 15.65 & 58.65 & 94.59 & 69.04
& 97.76 & 96.52 & 83.31 \\
Rolling-Sink~\citep{li2026rolling}
& \underline{16.45} & \underline{61.61} & 96.13 & \underline{71.89}
& \textbf{98.84} & 97.96 & \underline{85.29} \\
LongLive~\citep{yang2025longlive}
& 16.41 & 59.00 & 95.33 & 69.30
& 98.09 & 97.32 & 83.81 \\
Rolling-Forcing~\citep{liu2025rolling}
& 15.79 & 60.93 & 96.26 & 69.37
& 98.57 & \underline{97.99} & 84.62 \\
\rowcolor{SOTARow}
Ours
& 16.01 & \textbf{61.69} & \textbf{97.17} & \textbf{72.09}
& \underline{98.79} & \textbf{98.28} & \textbf{85.60} \\

\midrule
\multicolumn{8}{c}{\textbf{Results on 120s}} \\
\midrule
Self-Forcing~\citep{huang2025self}
& \textbf{17.01} & 50.37 & \textbf{97.15} & 61.36
& 95.17 & 98.07 & 80.42 \\
$\infty$-RoPE~\citep{yesiltepe2025infinity}
& \textbf{17.01} & 59.12 & 95.70 & 69.75
& 98.27 & 96.95 & 83.96 \\
Deep-Forcing~\citep{yi2025deep}
& 15.65 & 58.59 & 96.37 & 68.11
& 98.45 & 97.56 & 83.82 \\
Rolling-Sink~\citep{li2026rolling}
& \underline{16.45} & \underline{61.53} & 96.27 & \underline{70.48}
& \underline{98.99} & \underline{98.26} & \underline{85.11} \\
LongLive~\citep{yang2025longlive}
& 16.41 & 59.25 & 95.43 & 68.91
& 98.41 & 97.12 & 83.82 \\
Rolling-Forcing~\citep{liu2025rolling}
& 15.79 & 60.54 & 96.37 & 70.41
& 98.92 & 98.25 & 84.90 \\
\rowcolor{SOTARow}
Ours
& 16.01 & \textbf{61.75} & \underline{96.53} & \textbf{72.83}
& \textbf{99.05} & \textbf{98.33} & \textbf{85.70} \\
\bottomrule
\end{tabular}
\par}
\caption{
Evaluation of long-video generation on VBench-Long.
Comparison with multiple baselines at 60s and 120s across multiple VBench-Long metrics, with Echo-Forcing consistently achieving the best overall performance at both durations.
}
\label{tab:main_experiment_long}
\end{table*}


\begin{table*}[t]
\centering
{\small
\setlength{\tabcolsep}{5.5pt}
\renewcommand{\arraystretch}{0.95}
\begin{tabular}{llcccccc}
\toprule
& & \multicolumn{5}{c}{\textit{Segment-wise Matching Metrics}} & \\
\cmidrule(lr){3-7}
Mode & Method
& \begin{tabular}{@{}c@{}}Text\\ Align.${\mathbf{\red{\uparrow}}}$\end{tabular}
& \begin{tabular}{@{}c@{}}Subject\\ Consist.${\mathbf{\red{\uparrow}}}$\end{tabular}
& \begin{tabular}{@{}c@{}}Background\\ Consist.${\mathbf{\red{\uparrow}}}$\end{tabular}
& \begin{tabular}{@{}c@{}}Aesthetic\\ Quality${\mathbf{\red{\uparrow}}}$\end{tabular}
& \begin{tabular}{@{}c@{}}Imaging\\ Quality${\mathbf{\red{\uparrow}}}$\end{tabular}
& \begin{tabular}{@{}c@{}}Overall\\ Avg.${\mathbf{\red{\uparrow}}}$\end{tabular} \\
\midrule

\multicolumn{8}{c}{\textit{Unfine-tuned Models}} \\
\midrule
\multirow{3}{*}{Smooth Transition}
& Self-Forcing + Recache~\citep{yang2025longlive} & \underline{26.92} & \underline{86.13} & 86.48 & 50.15 & 50.22 & 59.98 \\
& $\infty$-RoPE + Flush~\citep{yesiltepe2025infinity} & 25.94 & 83.95 & \underline{87.94} & \underline{57.33} & \underline{66.81} & \underline{64.39} \\
\rowcolor{SOTARow}
& Echo-Forcing & \textbf{27.94} & \textbf{93.19} & \textbf{92.63} & \textbf{58.64} & \textbf{69.21} & \textbf{68.32} \\
\cmidrule(lr){1-8}

\multirow{3}{*}{Hard Cut}
& Self-Forcing + Recache~\citep{yang2025longlive} & 29.88 & 76.43 & \underline{86.77} & 50.02 & 50.17 & 58.65 \\
& $\infty$-RoPE + Flush~\citep{yesiltepe2025infinity} & \underline{32.63} & \underline{78.88} & 85.89 & \underline{57.12} & \underline{67.79} & \underline{64.46} \\
\rowcolor{SOTARow}
& Echo-Forcing & \textbf{33.67} & \textbf{79.39} & \textbf{89.62} & \textbf{58.00} & \textbf{69.64} & \textbf{66.06} \\
\cmidrule(lr){1-8}

\multirow{3}{*}{Scene Recall}
& Self-Forcing + Recache~\citep{yang2025longlive} & 28.58 & 78.12 & 74.71 & 50.98 & 51.02 & 56.68 \\
& $\infty$-RoPE + Flush~\citep{yesiltepe2025infinity} & \underline{29.47} & \underline{79.31} & \underline{78.39} & \underline{56.31} & \underline{66.73} & \underline{62.04} \\
\rowcolor{SOTARow}
& Echo-Forcing & \textbf{32.58} & \textbf{83.11} & \textbf{81.57} & \textbf{58.76} & \textbf{68.88} & \textbf{64.98} \\

\midrule
\multicolumn{8}{c}{\textit{Fine-tuned Models}} \\
\midrule
\multirow{3}{*}{Smooth Transition}

& LongLive + Recache~\citep{yang2025longlive} & \underline{27.38} & 94.01 & 91.39 & \textbf{54.89} & 67.17 & 66.97 \\
& LongLive + Flush~\citep{yesiltepe2025infinity} & 27.34 & \underline{94.22} & \underline{91.88} & \underline{54.85} & \underline{67.33} & \underline{67.12} \\
\rowcolor{SOTARow}
& LongLive + Ours & \textbf{29.77} & \textbf{95.32} & \textbf{93.74} & 54.78 & \textbf{67.87} & \textbf{68.30} \\
\cmidrule(lr){1-8}

\multirow{3}{*}{Hard Cut}

& LongLive + Recache~\citep{yang2025longlive} & 30.59 & 81.53 & \textbf{57.78} & 54.46 & 67.76 & 58.42 \\
& LongLive + Flush~\citep{yesiltepe2025infinity} & \underline{31.97} & \underline{81.99} & 56.33 & \underline{54.49} & \underline{69.01} & \underline{58.76} \\
\rowcolor{SOTARow}
& LongLive + Ours & \textbf{34.27} & \textbf{83.39} & \underline{57.75} & \textbf{55.01} & \textbf{69.77} & \textbf{60.04} \\
\cmidrule(lr){1-8}

\multirow{3}{*}{Scene Recall}
& LongLive + Recache~\citep{yang2025longlive} & 28.56 & 82.23 & \underline{86.50} & 53.31 & 66.32 & 63.38 \\
& LongLive + Flush~\citep{yesiltepe2025infinity} & \underline{30.18} & \underline{82.77} & 86.23 & \underline{53.77} & \underline{67.43} & \underline{64.08} \\
\rowcolor{SOTARow}
& LongLive + Ours & \textbf{32.58} & \textbf{83.48} & \textbf{87.57} & \textbf{54.98} & \textbf{69.13} & \textbf{65.55} \\
\bottomrule
\end{tabular}

\par}
\caption{
Evaluation of interactive video generation.
VBench-Long metrics and text alignment are computed after segment-wise matching across smooth transitions, hard cuts, and scene recall. Echo-Forcing achieves the highest Overall Avg. for each interaction mode in both model groups.
}
\label{tab:main_experiment_inter}
\end{table*}

\noindent\textbf{Overview.}
Echo-Forcing assigns its three modules to distinct stages of the interactive generation lifecycle: \textbf{Hierarchical Temporal Memory} supports memory maintenance during ongoing generation, \textbf{Scene Recall Frames} retrieve historical scene priors at scene transitions, and \textbf{Difference-aware Memory Decay} attenuates conflicting memories after each transition. Together, they enable stage-aware KV memory management throughout the full generation lifecycle.

\subsection{Hierarchical Temporal Memory}
\label{sec:hierarchical_temporal_memory}

Uniform sliding-window caching reuses noisy history and amplifies errors in autoregressive long-video generation. Historical KV states serve distinct roles across temporal scales: early, long-range, and recent contexts support stability, global evolution, and local continuity, respectively. As illustrated in Figure~\ref{fig:overview}, \textbf{Hierarchical Temporal Memory} separates the KV cache into complementary temporal memories using rolling anchors and phase-calibrated compression. We use relative RoPE extrapolation to bound temporal indices during rollout (Appendix~\ref{sec:relative-rope}).

\subsubsection{Bidirectional Rolling Early Anchors}
\label{sec:bidirectional_rolling_anchors}

Early frames are generated within the training horizon and thus provide relatively clean global references. We reuse them as early anchors for long-horizon generation. Let \(N_{\mathrm{anc}}\) denote the anchor-pool size, where \(E_i\) represents the raw KV tokens of the \(i\)-th anchor frame. At the \(r\)-th update, we insert \(S\) anchors starting from \(u_r=(rS)\bmod N_{\mathrm{anc}}\). The inserted sequence is defined as
\begin{equation}
\mathcal{A}_r=
\begin{cases}
(E_{u_r}, E_{u_r+1}, \ldots, E_{u_r+S-1}), & r\ \text{is odd},\\
(E_{u_r+S-1}, E_{u_r+S-2}, \ldots, E_{u_r}), & r\ \text{is even}.
\end{cases}
\label{eq:bidirectional_anchors}
\end{equation}
All indices are taken modulo \(N_{\mathrm{anc}}\). Consecutive updates traverse the anchor pool in alternating forward and backward orders, refreshing stable references while avoiding a fixed anchor ordering. After each update, \(\mathcal{A}_r\) is appended to the anchor memory, providing persistent early-stage references with negligible cache overhead.

\subsubsection{Drift-gated phase compression}
\label{sec:drift_gated_phase_compression}

To retain informative long-range tokens, we propose \textbf{Drift-Gated Phase Compression}. It estimates the future usefulness of historical tokens using a calibrated pre-RoPE query reference, and adaptively adjusts the estimate according to recent query drift. See Appendix~\ref{sec:compression} for details.

\paragraph{\textit{Online calibration.}}
We construct a stable phase reference from the pre-RoPE queries collected during the early calibration stage. Let \(\mathcal{Q}_{\mathrm{cal}}\) denote the calibration query set. We compute
\begin{equation}
\bar{\mathbf{q}}
= \frac{1}{|\mathcal{Q}_{\mathrm{cal}}|}\sum_{q \in \mathcal{Q}_{\mathrm{cal}}} \mathbf{q},
\qquad
\bar{a}_q
= \frac{1}{|\mathcal{Q}_{\mathrm{cal}}|}\sum_{q \in \mathcal{Q}_{\mathrm{cal}}} |\mathbf{q}|.
\end{equation}
Here, \(\bar {\mathbf{q}}\) provides a stable query direction for phase-coherent scoring, while \(\bar a_q\) records the typical query magnitude for amplitude compensation. Both are computed from the normal forward pass without extra inference cost.

\paragraph{\textit{Token importance scoring.}}
Following the trigonometric decomposition of RoPE attention in TriAttention~\cite{mao2026triattention}, we score historical pre-RoPE keys in the complex domain. Here, \(f\) denotes the RoPE frequency-channel index in the complex representation. For a historical token \(j\) with pre-RoPE key \(\mathbf{k}^{\mathrm{raw}}_j\), we define the per-channel phase gap as
\(\phi_{j,f}=\arg(\bar {\mathbf{q}}_f \overline{\mathbf{k}^{\mathrm{raw}}_{j,f}})\).
Given the next frame index \(a_b^+\), the token frame index \(a_j\), and a future offset \(o\in\mathcal{O}\), the temporal distance is
\(\Delta_{j,o}=a_b^+-a_j+o\). We compute the phase-coherent score as
\begin{equation}
\mathrm{Score}^{\mathrm{ph}}_{j,o}
=
\sum_f
|\bar {\mathbf{q}}_f|\,|\mathbf{k}^{\mathrm{raw}}_{j,f}|
\cos\left(\phi_{j,f}+\omega_f\Delta_{j,o}\right).
\end{equation}
This score estimates how well a historical token remains phase-aligned with future queries after RoPE temporal evolution, allowing selection to depend on expected future usefulness rather than immediate attention to the current block.

In addition, we compute a magnitude compensation term from the calibration statistics:
\begin{equation}
\mathrm{AMP}_j
=
\sum_f
(\bar a_{q,f}-|\bar {\mathbf{q}}_f|)
|\mathbf{k}^{\mathrm{raw}}_{j,f}|.
\end{equation}
This term captures the query magnitude component not represented by the dominant calibrated direction, and will be adaptively modulated by the drift gate in the final selection score.

\begin{figure}[h]
    \centering
    \includegraphics[width=1.0\linewidth]{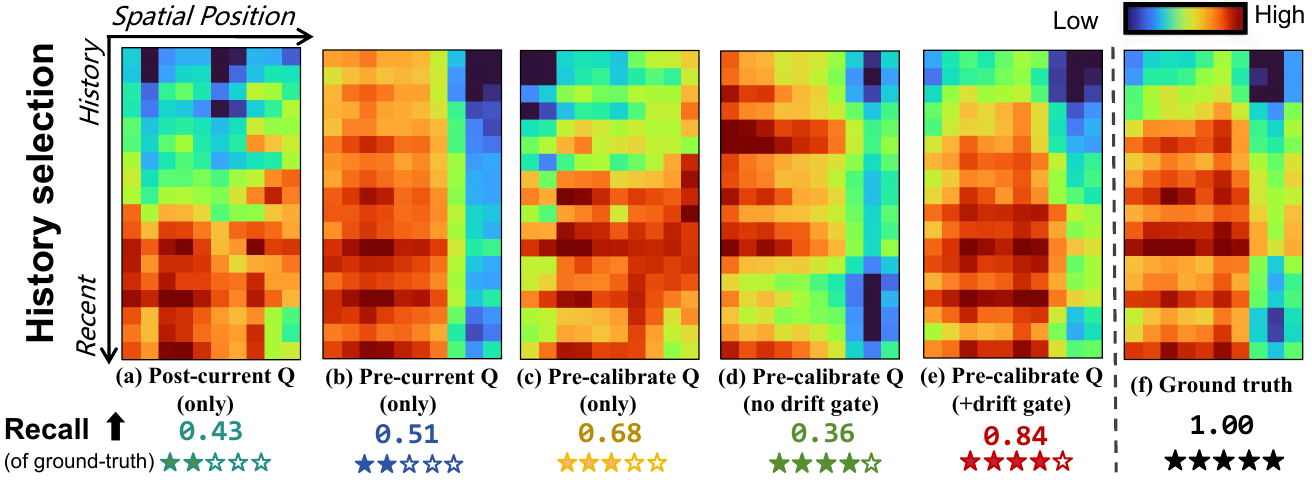}
    \caption{
    Comparison of historical token-selection strategies.
    Heatmaps show spatial token-selection weights, and recall measures agreement with the ground-truth future-query selection in (f). Pre-RoPE scoring, query calibration, and drift gating progressively improve this alignment.
    }
    \label{fig:token_selection}
\end{figure}

\paragraph{\textit{Drift Gate.}}
Although the calibration center provides a stable reference, recent queries may gradually deviate from the early calibrated distribution as long-horizon generation proceeds, making fixed magnitude compensation unreliable. We therefore introduce a drift gate based on the similarity between the recent query center \(\bar{\mathbf{q}}_{\mathrm{rec}}\) and the calibration center \(\bar{\mathbf{q}}\), which adaptively modulates magnitude compensation to avoid inappropriate amplification of historical memory under query drift:

\begin{equation}
\begin{aligned}
g_b
&=
\exp\!\left[
-\lambda\left(
1-\cos\left(
\bar{\mathbf{q}}_{\mathrm{rec}},
\bar{\mathbf{q}}
\right)
\right)
\right],
\\
\mathrm{Score}_j
&=
\operatorname{Fuse}_{o\in\mathcal{O}}
\left(
\mathrm{Score}^{\mathrm{ph}}_{j,o}
-
g_b\,\mathrm{AMP}_j
\right).
\end{aligned}
\end{equation}

\(\lambda\) denotes the drift-gate sensitivity coefficient. This gate adaptively controls how much the magnitude compensation contributes to the final selection score. When recent queries remain close to the calibration center, \(g_b\) keeps the compensation term active to capture useful dynamic response strength. When the drift becomes large, \(g_b\) suppresses the compensation term and makes the selection rely more on phase-coherent alignment, preventing unstable recent queries from over-amplifying noisy or mismatched historical tokens.

Finally, we exclude recent-window tokens \(\mathcal{R}_b\) from compression and retain the top-\(K\) historical tokens:

\begin{equation}
\mathcal{I}^{\mathrm{cmp}}_b
=
\operatorname{TopK}_{j\in\mathcal{H}_b\setminus\mathcal{R}_b}
\left(\mathrm{Score}_j\right).
\label{eq:compression_selection}
\end{equation}

As shown in Figure~\ref{fig:token_selection}, the proposed scoring better aligns token selection with ground-truth future-query attention, retaining phase-consistent long-range tokens without over-relying on stale calibration statistics or noisy recent queries.

\begin{figure*}[t]
    \centering

    \subfloat[
    \textbf{Scene Recall Frames.}
    Spatial aggregation retains scene cues and better follows target prompts than key-frame selection.
    ]{
        \includegraphics[width=0.92\textwidth]{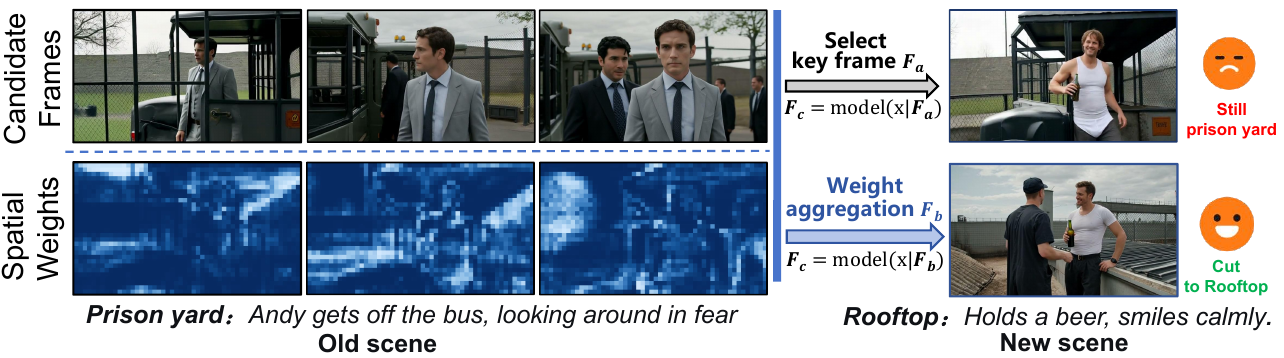}
        \label{fig:scene_recall_insight}
    }

    \subfloat[
    \textbf{Difference-aware Memory Decay.}
    Position-wise decay preserves compatible regions and forgets conflicts.
    ]{
        \includegraphics[width=0.92\textwidth]{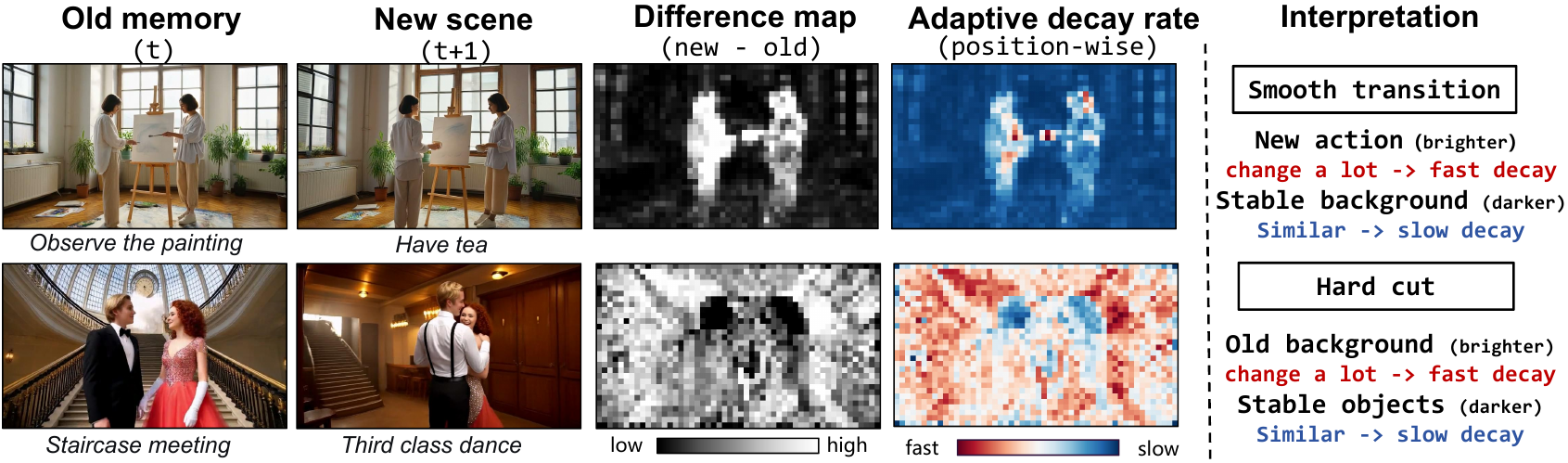}
        \label{fig:decay_insight}
    }

    \caption{
    Visualization of scene-memory management.
    Echo-Forcing combines compact scene recall with position-wise memory decay to retrieve relevant historical priors and suppress conflicting memory during interactive generation.
    }
    \label{fig:scene_memory}
\end{figure*}

\subsection{Scene Recall Frames}
\label{sec:scene_recall_frames}

Existing interactive autoregressive methods largely focus on transitions between adjacent scenes and do not maintain explicit scene-level memories, making long-range scene recall difficult. Storing all KV states from a historical scene is costly and introduces redundant interference, whereas a single frame cannot adequately represent its temporal variations. We therefore propose \textbf{Scene Recall Frames}, which aggregate multi-frame KV tokens into compact scene representations. At scene transitions, the corresponding representation can be retrieved as a scene prior for generation. Additional details are provided in Appendix~\ref{sec:recall}.

For the \(s\)-th scene, we collect \(M\) candidate blocks from its stable generation stage, each retaining the full spatial token layout. For spatial position \(u\), we compute the aggregation weight of candidate block \(j\) using the calibrated query center \(\bar{\mathbf{q}}_{s,u}\):
\begin{equation}
\alpha_{s,j,u}
=
\operatorname{Softmax}_{j}
\left(
\operatorname{sim}
\left(
\bar{\mathbf{q}}_{s,u},
\mathbf{k}^{\mathrm{raw}}_{s,j,u}
\right)
\right).
\label{eq:scene_recall_weight}
\end{equation}
The recall KV tokens are obtained by spatially weighted fusion:
\begin{equation}
\left(
\mathbf{K}^{\mathrm{rec}}_{s,u},
\mathbf{V}^{\mathrm{rec}}_{s,u}
\right)
=
\sum_{j=1}^{M}
\alpha_{s,j,u}
\left(
\mathbf{K}^{\mathrm{raw}}_{s,j,u},
\mathbf{V}_{s,j,u}
\right).
\label{eq:scene_recall_fusion}
\end{equation}

The resulting recall frames are stored in a scene-memory pool and retrieved when the corresponding scene is requested. As shown in Figure~\ref{fig:scene_memory}(a), spatially weighted aggregation preserves prompt-relevant scene cues more effectively than key frame selection, while retaining complementary information across multiple frames with substantially lower cache overhead than full-scene storage.

\subsection{Difference-aware Memory Decay}
After a scene transition, KV memories preserve transferable priors, such as subject identity and scene structure, but may conflict with the new prompt. Frame-wise cache operations treat all tokens uniformly and cannot distinguish compatible regions from obsolete ones. We propose \textbf{Difference-aware Memory Decay}, which assigns position-wise token decay rates based on old--new scene discrepancies, retaining compatible memories while progressively suppressing conflicting regions during post-transition generation.

\paragraph{Discrepancy-aware Estimation.}
After entering a new scene, we first generate its initial clean block as a new-scene reference. Let \((\mathbf{k}_i^{\mathrm{old}},\mathbf{v}_i^{\mathrm{old}})\) denote an old-memory token, and let \(\mathbf{k}_i^{\mathrm{new}}\) be the key at its corresponding or neighboring spatial position in the reference block. We measure their feature discrepancy as
\begin{equation}
d_i
=
1-\cos\!\left(
\mathbf{k}_i^{\mathrm{old}},
\mathbf{k}_i^{\mathrm{new}}
\right).
\end{equation}

We then map the normalized discrepancy to a token-wise decay rate:
\begin{equation}
\mu_i
=
\mu_{\min}
+
(\mu_{\max}-\mu_{\min})
\operatorname{Norm}(d_i),
\label{eq:decay_rate}
\end{equation}
where \(\operatorname{Norm}(\cdot)\) denotes normalization across old-memory tokens. A larger discrepancy \(d_i\) produces a larger decay rate \(\mu_i\), allowing changed regions to be forgotten faster while preserving compatible memories longer.

\paragraph{KV-Level Soft Forgetting.}
For each old token, we maintain a memory weight \(w_i^{(r)}\), where \(r\) denotes the generation step after the transition. The weight is initialized as \(w_i^{(0)}=1\) and decays exponentially:

\begin{equation}
\begin{aligned}
w_i^{(r)}
&=
\exp(-r\mu_i),
\\[-1mm]
\left(
\tilde{\mathbf{k}}_i^{(r)},
\tilde{\mathbf{v}}_i^{(r)}
\right)
&=
w_i^{(r)}
\left(
\mathbf{k}_i^{\mathrm{old}},
\mathbf{v}_i^{\mathrm{old}}
\right).
\end{aligned}
\end{equation}

Applying the decay to both keys and values suppresses the old token in attention matching and weakens its contribution to the output:

\begin{equation}
\begin{aligned}
\ell_{q,i}^{(r)}
&=
\frac{\mathbf{q}^{\top}\tilde{\mathbf{k}}_i^{(r)}}{\sqrt{d}}
=
w_i^{(r)}
\frac{\mathbf{q}^{\top}\mathbf{k}_i^{\mathrm{old}}}{\sqrt{d}},
\\[-1mm]
o_q^{(r)}
&=
\sum_i
\operatorname{Softmax}_{i}
\left(\ell_{q,i}^{(r)}\right)
\tilde{\mathbf{v}}_i^{(r)}.
\end{aligned}
\end{equation}

In this way, compatible old memories can still support the early transition, while conflicting regions are rapidly suppressed, allowing the new scene to gradually dominate the generation process. See Appendix~\ref{sec:decay} for details.

\begin{figure*}[t]
    \centering
    \includegraphics[width=1.0\textwidth]{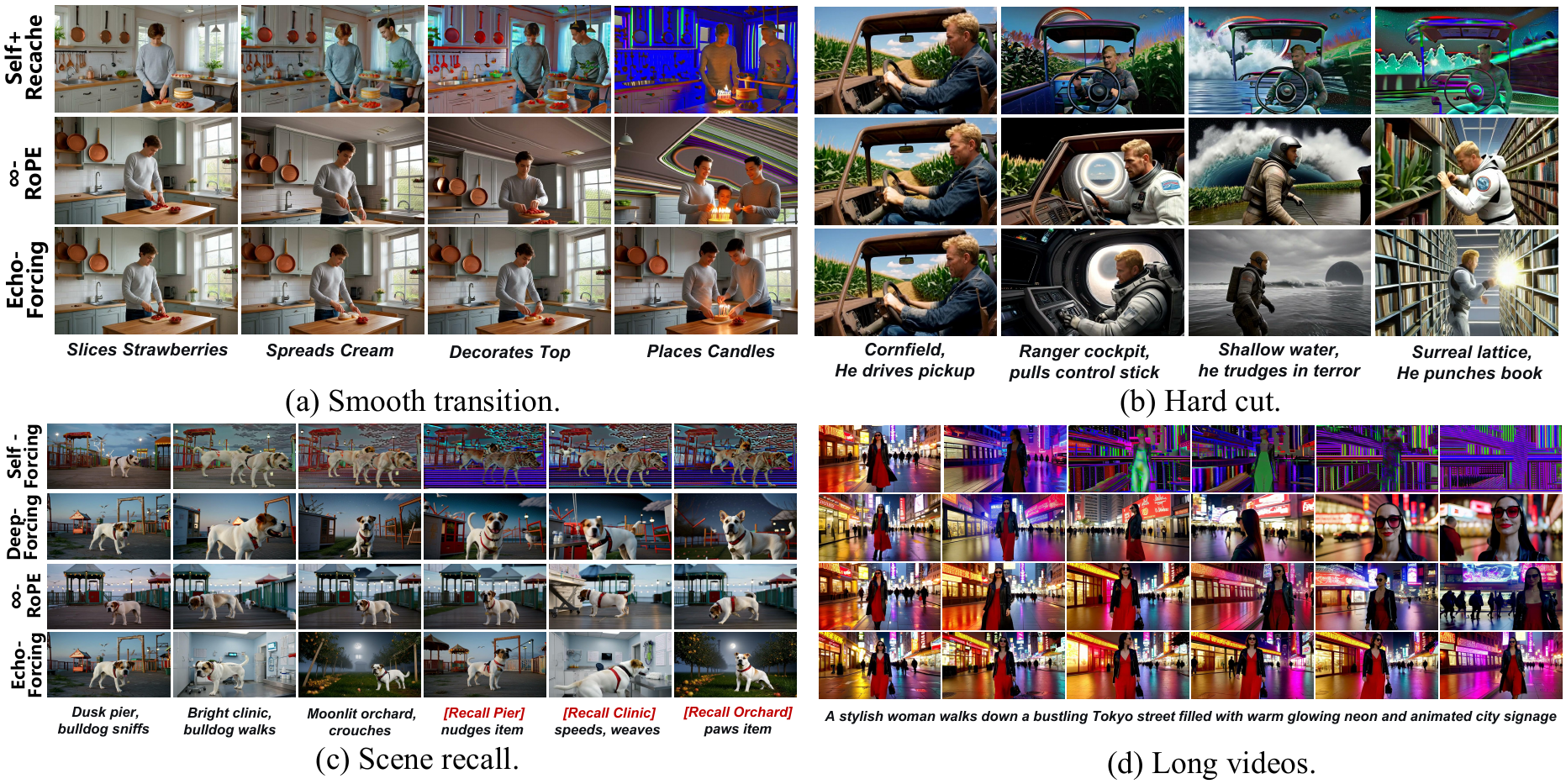}
    \caption{
    Qualitative comparison.
    Across smooth transitions, hard cuts, scene recall, and long-video generation, Echo-Forcing, compared with prior methods, better follows evolving prompts while preserving subject and visual coherence over long horizons.
    }
    \label{fig:compare}
\end{figure*}

\section{Experiments}

\subsection{Experimental setup}

\paragraph{Implementation details.}
For fair comparison, all methods use a window of \(L=21\) frames. We use chunk-wise Self-Forcing~\citep{huang2025self} and LongLive~\citep{yang2025longlive} as the non-fine-tuned and fine-tuned bases, respectively. Echo-Forcing uses \(N_{\mathrm{anc}}=12\) rolling anchors, \(N_{\mathrm{cmp}}=3\) compressed history frames, and \(N_{\mathrm{rec}}=3\) recent frames with relative-time RoPE with 3 frames for current generation. For scene recall, the recall-frame size is 1 and the pool capacity is 64. All experiments are conducted on NVIDIA H100 GPUs. Additional implementation details and automatic scene routing are provided in Appendices~\ref{app:ablation} and~\ref{app:routing}.

\paragraph{Benchmarks.}
Following $\infty$-RoPE, we use VBench-Long~\citep{huang2024vbench,zheng2025vbench2,huang2025vbench++} to evaluate long-video and interactive generation. In both settings, prompts are sampled from MovieGenBench~\citep{polyak2024moviegen} and expanded using Qwen/Qwen2.5-7B-Instruct~\citep{hui2024qwen2}, following Self-Forcing~\citep{huang2025self}. For long-video generation, we sample 128/64 prompts for 60s/120s videos. For interactive generation, we construct smooth-transition, hard-cut, and scene-recall subsets, each containing 64 six-shot 60s samples. Videos are segmented by scenes for paired evaluation, with scene orders randomly permuted. Further details are provided in Appendix~\ref{app:exp_details}.

\begin{table}[htbp]
\centering
{\small
\setlength{\tabcolsep}{3.2pt}
\renewcommand{\arraystretch}{1.0}
\begin{tabular}{lcccc}
\toprule
Method
& \begin{tabular}{@{}c@{}}Text\\ Align.${\mathbf{\red{\uparrow}}}$\end{tabular}
& \begin{tabular}{@{}c@{}}Subject\\ Consist.${\mathbf{\red{\uparrow}}}$\end{tabular}
& \begin{tabular}{@{}c@{}}Aesthetic\\ Quality${\mathbf{\red{\uparrow}}}$\end{tabular}
& \begin{tabular}{@{}c@{}}Recall\\ Acc.${\mathbf{\red{\uparrow}}}$\end{tabular} \\
\midrule
$\infty$-RoPE + Flush & 29.47 & 79.31 & 56.31 & -- \\
SwitchCraft  & 25.51 & 77.53 & 57.01 & -- \\
\rowcolor{SOTARow}
Echo-Forcing          & \textbf{32.58} & \textbf{83.11} & \textbf{58.76} & \textbf{99.48 \%} \\
\bottomrule
\end{tabular}
\par}
\caption{
Scene recall comparison.
We compare Echo-Forcing with Flush and SwitchCraft~\citep{xu2026switchcraft}, an interactive bidirectional baseline. All metrics use segment-wise matching, with Recall Acc. measuring historical-scene retrieval.
}
\label{tab:wmq_recall_accuracy}
\end{table}

\paragraph{Quantitative and qualitative results.}
Tables~\ref{tab:main_experiment_long} and~\ref{tab:main_experiment_inter} compare Echo-Forcing with non-fine-tuned and fine-tuned baselines. For long-video generation, Echo-Forcing achieves the best overall scores at 60s/120s (\textbf{85.60}/\textbf{85.70}) and the highest imaging quality (\textbf{72.09}/\textbf{72.83}). Across interactive modes, it also achieves the best text alignment and overall scores; for scene recall, these reach \textbf{32.58}/\textbf{64.98} without fine-tuning and \textbf{32.58}/\textbf{65.55} with LongLive. Table~\ref{tab:wmq_recall_accuracy} further includes SwitchCraft~\citep{xu2026switchcraft}, a bidirectional interactive baseline implemented on Self-Forcing and evaluated via Recall Acc., Echo-Forcing achieves \textbf{99.48 \% }, while other interactive methods lack explicit scene-retrieval capability. See Appendix~\ref{app:routing} for details. Figure~\ref{fig:compare} shows more coherent long videos and more reliable responses to interactive scene changes. Additional qualitative results are provided in Appendix~\ref{app:visualizations}, Figures~\ref{fig:app_add_long}--\ref{fig:app_add_mem}.

\subsection{Ablation studies}
\label{subsec:ablation}

We conduct ablations on both long-video memory organization and interactive scene-memory management, covering rolling-anchor update mode (Table~\ref{tab:rolling_mode}), memory-budget allocation (Table~\ref{tab:ablation_compress_size}), drift-gated phase compression (Table~\ref{tab:ablation_drift_gate}), drift-gate coefficient sensitivity (Table~\ref{tab:ablation_gate_value}), and interactive scene-memory design, including scene recall source and memory decay strategy (Table~\ref{tab:ablation_recall_decay}).
We present the key ablation on Drift-Gated Phase Compression in the main paper, while the remaining studies are provided in Appendix~\ref{app:ablation}.

\begin{table}[htbp]
\centering
{\small
\setlength{\tabcolsep}{3.2pt}
\renewcommand{\arraystretch}{1.0}

\begin{tabular}{ccc|cccc}
\toprule
\begin{tabular}{@{}c@{}}Phase\\ Score\end{tabular}
& AMP
& \begin{tabular}{@{}c@{}}Drift\\ Gate\end{tabular}
& \begin{tabular}{@{}c@{}}Aesthetic\\ Quality${\mathbf{\red{\uparrow}}}$\end{tabular}
& \begin{tabular}{@{}c@{}}Subject\\ Consist.${\mathbf{\red{\uparrow}}}$\end{tabular}
& \begin{tabular}{@{}c@{}}Dynamic\\ Degree${\mathbf{\red{\uparrow}}}$\end{tabular}
& \begin{tabular}{@{}c@{}}Overall\\ Avg.${\mathbf{\red{\uparrow}}}$\end{tabular} \\
\midrule
\cmark & \xmark & \xmark & 61.24 & 97.00 & 35.31 & 64.52 \\
\xmark & \cmark & \xmark & 59.34 & 96.11 & 30.51 & 61.99 \\
\cmark & \cmark & \xmark & \underline{61.67} & \underline{97.16} & \underline{37.08} & \underline{65.30} \\
\cmark & \cmark & \cmark & \textbf{61.69} & \textbf{97.17} & \textbf{47.59} & \textbf{68.82} \\
\bottomrule
\end{tabular}
\par}
\caption{
Ablation of Drift-Gated Phase Compression.
We isolate phase-coherent scoring, amplitude compensation (AMP), and drift gating. Drift gating adaptively regulates AMP under query-distribution shifts.
}
\label{tab:ablation_drift_gate}
\end{table}

\paragraph{Drift-Gated Phase Compression.}
Table~\ref{tab:ablation_drift_gate} validates the contributions of phase-coherent scoring, amplitude compensation, and drift gating. The full design achieves the best aesthetic quality and subject consistency, resulting in the highest overall score of \textbf{68.82}, confirming that drift gating enables reliable historical-token selection under query drift.

\subsection{User Studies}
\label{subsec:user_studies}

\begin{table}[t]
\centering

\begin{minipage}[t]{0.98\linewidth}
\centering
{\small
\setlength{\tabcolsep}{0.35pt}
\renewcommand{\arraystretch}{1.0}
\begin{tabular}{l|ccccc}
\toprule
Method
& \begin{tabular}{@{}c@{}}Prompt\\ Follow$\textcolor{red}{\uparrow}$\end{tabular}
& \begin{tabular}{@{}c@{}}Subject\\ Consist.$\textcolor{red}{\uparrow}$\end{tabular}
& \begin{tabular}{@{}c@{}}Motion\\ Quality.$\textcolor{red}{\uparrow}$\end{tabular}
& \begin{tabular}{@{}c@{}}Video\\ Quality$\textcolor{red}{\uparrow}$\end{tabular}
& \begin{tabular}{@{}c@{}}Overall\\ Avg.$\textcolor{red}{\uparrow}$\end{tabular}
\\
\midrule

Self-Forcing
& 2.44 & 2.81 & 2.69 & 2.24
& 2.55
\\
Deep-Forcing
& 2.89 & 2.43 & 2.03 & 1.13
& 2.12
\\

$\infty$-RoPE
& 2.97 & 2.83 & 3.05 & 3.04
& 2.97
\\

Rolling-Forcing
& 3.10
& 2.98
& 2.78
& 3.13
& 3.00
\\

LongLive
& \underline{3.16} & \underline{3.23} & \underline{3.08} & \underline{3.34}
& \underline{3.20}
\\
\rowcolor{SOTARow}
Ours
& \textbf{3.52}
& \textbf{3.33}
& \textbf{3.64}
& \textbf{3.41}
& \textbf{3.48}
\\

\bottomrule
\end{tabular}
\par}
\captionof{table}{
User study for long-video generation.
Evaluating 24 120-second videos.
}
\label{tab:user_study_long}
\end{minipage}

\par

\begin{minipage}[t]{0.98\linewidth}
\centering
{\small
\setlength{\tabcolsep}{0.35pt}
\renewcommand{\arraystretch}{1.0}
\begin{tabular}{l|ccccc}
\toprule
Method
& \begin{tabular}{@{}c@{}}Instruction\\ Follow$\textcolor{red}{\uparrow}$\end{tabular}
& \begin{tabular}{@{}c@{}}Subject\\ Consist.$\textcolor{red}{\uparrow}$\end{tabular}
& \begin{tabular}{@{}c@{}}Motion\\ Quality.$\textcolor{red}{\uparrow}$\end{tabular}
& \begin{tabular}{@{}c@{}}Video\\ Quality$\textcolor{red}{\uparrow}$\end{tabular}
& \begin{tabular}{@{}c@{}}Overall\\ Avg.$\textcolor{red}{\uparrow}$\end{tabular}
\\
\midrule

Self-Forcing
& 2.51
& 2.36
& 2.21
& 2.01
& 2.27
\\

$\infty$-RoPE
& \underline{3.59}
& \underline{3.42}
& 3.57
& \underline{3.53}
& \underline{3.53}
\\

LongLive
& 3.16
& 3.18
& \underline{3.61}
& 3.23
& 3.30
\\

\rowcolor{SOTARow}
Ours
& \textbf{4.50}
& \textbf{4.27}
& \textbf{3.91}
& \textbf{3.77}
& \textbf{4.11}
\\

\bottomrule
\end{tabular}
\par}
\captionof{table}{
User study for interactive video generation.
Evaluating 36 60-second interactive videos with six scenes.
}
\label{tab:user_study_inter}
\end{minipage}
\end{table}

Following the five-point Likert-scale human-evaluation protocol of $\infty$-RoPE~\citep{yesiltepe2025infinity}, we conduct two blind user studies with 24 participants for long-video and interactive video generation. Participants rate anonymized videos in randomized order on 
prompt/instruction following, subject consistency, motion quality, and video quality. We report mean opinion scores over all valid ratings. As shown in Tables~\ref{tab:user_study_long} and~\ref{tab:user_study_inter}, Echo-Forcing achieves the highest overall scores of 3.48 for long-video generation and 4.11 for interactive generation. It ranks first on all dimensions in both evaluations. These results agree with the automatic metrics; details are provided in Appendix~\ref{sec:user}.

\section{Conclusion}
We presented \textbf{Echo-Forcing}, a training-free framework for lifecycle-aware KV memory management in interactive autoregressive video generation. Echo-Forcing treats historical KV states as evolving memories: they are structured and updated during generation, retrieved at scene transitions, and adaptively retained or decayed afterwards. By coordinating these stages under a bounded cache budget, it unifies stable long-video extension and flexible multi-scene interaction in one inference process. Experiments on VBench-Long and interactive generation settings demonstrate improved long-horizon quality, scene consistency, and responsiveness to evolving instructions without fine-tuning the video diffusion model. We hope Echo-Forcing provides a foundation for flexible and controllable streaming video generation.

\section*{Acknowledgments}
This work was supported by the National Natural Science Foundation of China under Grant Nos. 62676389, 62406312, 62476264, 62606510, the Youth Key Deployment Project of the Chinese Academy of Sciences under Grant No. \mbox{GFQN-2026-34}, and the Beijing Natural Science Foundation under Grant No. QG26011.

\title{Supplementary Material: Echo-Forcing: Lifecycle-Aware KV Memory Management for Interactive Autoregressive Long-Video Generation}

\bibliography{main}

\clearpage

\appendix
\setcounter{secnumdepth}{2}

\twocolumn[
    \begin{center}
        {\LARGE\bfseries Supplementary Material\par}
        \vspace{0.8em}
        {\LARGE\bfseries
        Echo-Forcing: Lifecycle-Aware KV Memory Management\\
        for Interactive Autoregressive Long-Video Generation\par}
        \vspace{1em}
    \end{center}
]

\renewcommand{\thetable}{\arabic{table}}
\renewcommand{\thefigure}{\arabic{figure}}
\renewcommand{\theequation}{\arabic{equation}}

\section{Dataset and evaluation details}
\label{app:exp_details}

\subsection{Dataset construction}
\label{sec:dataset}
Our evaluation data are built from prompts sampled from MovieGenBench~\citep{polyak2024moviegen}, following the long-video evaluation setups of Self-Forcing~\citep{huang2025self} and $\infty$-RoPE~\citep{yesiltepe2025infinity}.  We use Qwen/Qwen2.5-7B-Instruct~\citep{hui2024qwen2} to expand every base prompt into a detailed video description; the same expansion procedure is used for both the main paper and this appendix.

For long-video generation, we evaluate 60-second and 120-second rollouts.  We randomly sample 128 prompts for the 60-second setting and 64 prompts for the 120-second setting.  Each prompt is generated with four random seeds, and results are averaged over the resulting videos.  These prompt counts, durations, and generation settings are identical to those reported in the main paper.

For interactive generation, we construct three dedicated subsets for smooth transition, hard cut, and long-range scene recall.  Each subset contains 64 six-shot prompt sequences.  Each shot lasts 10 seconds, yielding a 60-second video.  After the scene specifications are constructed, we randomize their assignment to evaluation sequences and randomize the order of sequences and seeds during evaluation.  This prevents a method from benefiting from a fixed canonical scene order while retaining the temporal constraints required by each interaction mode.  All methods are evaluated on exactly the same randomized prompt sequences.

\paragraph{Smooth transition.}
In the smooth-transition subset, all six shots share the same subject identity, background context, and lighting style.  The transitions involve continuous actions, gradual pose changes, or smooth camera/viewpoint variations.  We randomize the selection and placement of the constructed scene variations before generation, while preserving their progressive semantic relation.  This setting evaluates whether a model can evolve a video smoothly without disrupting its established scene context.

\paragraph{Hard cut.}
In the hard-cut subset, the subject identity is preserved across all shots, whereas background, spatial layout, illumination, and action change substantially between consecutive shots.  We randomly permute the constructed cut scenes for each evaluation sequence, so the result is not tied to one particular succession of backgrounds.  This setting tests whether a model can respond to large prompt shifts while maintaining the appearance of the main subject, rather than leaking residual semantics from the previous scene.

\paragraph{Long-range scene recall.}
In the long-range scene-recall subset, the first three shots introduce three distinct scenes, denoted by $A$, $B$, and $C$.  The last three shots recall these scenes after a long temporal interval, while changing the camera viewpoint, subject pose, or action.  Rather than evaluating only the canonical $A$-$B$-$C$-$A$-$B$-$C$ order, we use all six recall orders: $ABCABC$, $ABCACB$, $ABCBAC$, $ABCBCA$, $ABCCAB$, and $ABCCBA$.  The 64 sequences are distributed as evenly as possible across these six forms and are randomized before evaluation.  Thus, recall performance cannot depend on a particular target position or recall order; it must retrieve the scene specified by the current prompt.

\subsection{Evaluation protocol}

\paragraph{Long-video generation.}
We evaluate 60-second and 120-second long videos with VBench-Long~\citep{huang2024vbench,zheng2025vbench2,huang2025vbench++}.  Following $\infty$-RoPE and LongLive, all VBench-Long metrics are computed on the complete generated video.  This full-video protocol measures long-horizon stability rather than the quality of isolated excerpts.

\paragraph{Segment matching for interactive generation.}
Interactive videos should not be evaluated as a single homogeneous VBench-Long clip: a single score can conflate intentional scene changes with generation errors.  We therefore use \emph{segment matching}.  Each 60-second output is split into six aligned 10-second segments, and segment $r$ is paired with its shot-level prompt $c_r$.  Per-segment scores are averaged only for metrics whose intended semantics are local.  The following mode-specific rules are used for every method.

\paragraph{Smooth transition.}
Text alignment is computed independently for each segment and its corresponding prompt, then averaged over the six segments.  Because the subject, background, and visual style are meant to remain continuous, subject consistency, background consistency, and video-quality metrics are computed on the complete 60-second video.  This avoids treating a gradual action or camera change as a discontinuity.

\paragraph{Hard cut.}
Text alignment is again evaluated segment by segment.  Subject and background consistency are computed at the complete-video scope to measure whether the desired global subject/background behavior is maintained across the instructed cuts.  In contrast, local visual-quality metrics (aesthetic quality, imaging quality, motion smoothness, temporal flickering, and dynamic degree) are computed separately on every segment and then averaged.  This prevents an intentional hard cut from being incorrectly penalized as a temporal artifact.

\paragraph{Long-range scene recall.}
Text alignment and local visual-quality metrics are computed for each segment and averaged.  For subject and background consistency, we first match every recalled segment to the earlier segment depicting the same scene, according to its randomized recall order.  We then compute the two consistency metrics on these matched segment pairs and average over all recalls.  This measures whether the requested historical scene is recovered, rather than rewarding similarity to the immediately preceding but irrelevant scene.

Overall, this segment-matching protocol preserves the VBench-Long metric definitions while assigning each metric a temporal scope consistent with the interactive instruction.

\subsection{User Study Protocol}
\label{sec:user}

Following the blind human-evaluation protocol of $\infty$-RoPE~\citep{yesiltepe2025infinity}, we conduct two studies with 24 participants with normal or corrected-to-normal vision.  We compare the training-free baselines and fine-tuned baseline reported in the main paper; the compared set includes Self-Forcing, Deep-Forcing, $\infty$-RoPE, Rolling-Forcing, LongLive, and the applicable Echo-Forcing variant.

For long-video generation, we randomly sample 24 MovieGenBench prompts and generate 120-second videos.  For interactive generation, we evaluate 36 six-shot, 60-second sequences: 12 smooth-transition, 12 hard-cut, and 12 scene-recall examples.  The chronological order within a video is retained so that motion and transitions can be judged, but prompt examples, methods, and display locations are randomized.  For each prompt, anonymized outputs from all compared methods are displayed simultaneously.  Method-to-slot assignments are independently shuffled and balanced across prompts, and prompts are presented in a randomized order.  This paired, simultaneous presentation follows $\infty$-RoPE and avoids bias caused by watching one method earlier, later, or consistently in a preferred screen position.

Participants score each anonymized output on a five-point Likert scale: 1 (very poor), 2 (poor), 3 (fair), 4 (good), and 5 (excellent).  For long videos, they rate prompt following, subject consistency across the 120-second rollout, motion quality (smoothness and naturalness), and overall visual quality (fidelity, artifacts, and aesthetics).  For interactive videos, they rate instruction following for all six shot prompts, subject consistency across transitions, transition motion quality, and overall visual quality.  The instructions additionally ask participants to judge gradual continuity for smooth transitions, responsiveness without stale-scene leakage for hard cuts, and agreement between each recalled segment and its matched earlier scene for recall.

We report mean opinion scores over valid ratings.  Echo-Forcing obtains the highest overall score in both studies: 3.48 for long-video generation and 4.11 for interactive generation.  On long videos, it improves on the strongest baseline (LongLive) by 0.28 overall, with the largest margin in motion quality (+0.56).  On interactive videos, it improves on the strongest overall baseline ($\infty$-RoPE) by 0.58, including +0.91 for instruction following and +0.85 for subject consistency.  These human preferences agree with the automatic results.

\section{Automatic Scene Switching and Routing Mechanism}
\label{app:routing}

The routing mechanism is optional.  Users can explicitly specify the interaction type by appending a control tag to a scene prompt, e.g., \texttt{[10s]} for smooth transition, \texttt{[10s\#]} for hard cut, and \texttt{[10s@]} for long-range scene recall.  A valid manual tag always takes precedence.  When no tag is provided, Echo-Forcing uses the following encoder-space routing rule; it does not rely on surface word overlap between raw prompts.

\paragraph{Prompt representation.}
Let $c_t$ be the current scene prompt.  We first remove its optional duration/control tag and tokenize the remaining natural-language prompt with the tokenizer of the frozen text encoder used by the video generator.  Let $\mathbf{e}_{t,m}$ be the contextual embedding of non-padding token $m$, and let $\mathcal{V}_t$ denote the valid-token set.  We form an $\ell_2$-normalized prompt representation by masked mean pooling:

\begin{equation}
\begin{gathered}
\mathbf{e}_{t,1:|\mathcal{V}_t|}
=
\Phi\!\left(\operatorname{Tok}(c_t)\right),\\
\mathbf{p}_t
=
\operatorname{Norm}_2\!\left(
\frac{1}{|\mathcal{V}_t|}
\sum_{m\in\mathcal{V}_t}\mathbf{e}_{t,m}
\right).
\end{gathered}
\label{eq:routing_prompt_embedding}
\end{equation}

Here, $\operatorname{Tok}(\cdot)$ is the encoder tokenizer, $\Phi(\cdot)$ is the frozen text encoder, and $\operatorname{Norm}_2(\cdot)$ denotes $\ell_2$ normalization.  Thus, prompts with different wordings but similar encoder representations can still be routed to the same historical scene.

For example, the prompt \emph{``A woman in a red coat walks on a neon Tokyo street [10s@]''} is first stripped to its natural-language content, tokenized by the encoder tokenizer, embedded token by token, and pooled into $\mathbf{p}_t$.  It is then compared with every earlier pooled prompt representation, rather than with the literal string.  If its highest similarity is to an earlier ``red-coat woman on Tokyo street'' scene rather than to the immediately preceding scene, the router selects that historical scene for recall.  This example illustrates the computation; the actual tokenizer may split words into subword units.

Let $\{\mathbf{p}_1,\dots,\mathbf{p}_{t-1}\}$ be the representations of previous scene prompts.  For the first scene, no routing is needed and it is treated as the initial context.  For $t>1$, we compute encoder-space cosine similarity between the current prompt and each previous prompt:
\begin{equation}
s_i = \cos(\mathbf{p}_t,\mathbf{p}_i)
=
\frac{\mathbf{p}_t^\top \mathbf{p}_i}{\|\mathbf{p}_t\|_2\,\|\mathbf{p}_i\|_2},
\quad i=1,\dots,t-1.
\end{equation}
We then identify the most similar historical scene:
\begin{equation}
s_{\max}=\max_{1\le i<t} s_i,
\qquad 
i^*=\arg\max_{1\le i<t} s_i.
\end{equation}

The transition mode $m_t$ is determined as
\begin{equation}
m_t =
\begin{cases}
\text{smooth}, & i^*=t-1 \ \text{and}\ s_{\max}\ge \tau_{\mathrm{smooth}},\\
\text{recall}, & i^*\neq t-1 \ \text{and}\ s_{\max}\ge \tau_{\mathrm{rec}},\\
\text{hard},   & \text{otherwise}.
\end{cases}
\end{equation}
Here, $\tau_{\mathrm{smooth}}$ controls whether the current scene should be regarded as a smooth continuation of the immediately preceding scene, while $\tau_{\mathrm{rec}}$ determines whether the current prompt is sufficiently similar to an earlier scene to trigger long-range recall. 
In our experiments, we set $\tau_{\mathrm{smooth}}=0.85$ and $\tau_{\mathrm{rec}}=0.85$.

We further assign a RoPE temporal offset $\Delta_t$ according to the inferred transition mode:
\begin{equation}
\Delta_t =
\begin{cases}
0, & m_t=\text{smooth},\\
45, & m_t=\text{hard},\\
\min(45,\gamma(t-i^*)), & m_t=\text{recall}.
\end{cases}
\end{equation}
where $\gamma=10$ controls how the offset increases with the temporal distance between the current scene and the recalled scene. 
This design keeps smooth transitions temporally continuous, introduces sufficient temporal separation for hard cuts, and assigns larger offsets to more distant recall targets while avoiding excessive positional extrapolation.

\paragraph{Recall-routing accuracy.}
The Recall Acc. reported in the main paper measures whether the automatic router retrieves the correct historical \emph{scene index}; it is not a visual-similarity score. We disable manual control tags, apply the encoder-space router to each recall query, and take the target scene specified by the prompt as ground truth. For $N_{\mathrm{rec}}$ recall queries, the accuracy is
\begin{equation}
\operatorname{Recall\ Acc.}
=
\frac{1}{N_{\mathrm{rec}}}
\sum_{n=1}^{N_{\mathrm{rec}}}
\mathbb{I}\!\left[\hat{i}_n=i_n^{\star}\right],
\label{eq:recall_routing_accuracy}
\end{equation}
where $\hat{i}_n=\arg\max_{i<t}s_i$ is the retrieved scene index, $i_n^{\star}$ is the annotated scene requested by the $n$-th prompt, and $\mathbb{I}[\cdot]$ is the indicator function. The recall benchmark contains six complete scene-order templates with different historical targets and recall positions, rather than a single fixed sequence. Therefore, this metric evaluates whether the router identifies the scene specified by the current prompt independently of its position in the sequence. Echo-Forcing obtains 99.48\% Recall Acc., showing that encoder-space matching identifies the requested history reliably under randomized orders.

\begin{samepage}
\subsection{Ambiguous prompt routing.}
Semantic routing cannot resolve every ambiguous instruction. A prompt may omit distinguishing details or describe multiple historical scenes. Echo-Forcing retrieves a scene by comparing the current prompt embedding with historical prompt embeddings. To evaluate robustness to reduced textual similarity, we test shortened, incomplete, and semantically overlapping prompts separately and in a mixed setting that combines all three.
We repeat the evaluation across multiple trials using a set of 192 recall queries. We report the \emph{top-1 scene hit rate}, counting a query as a hit when the highest-scoring historical scene matches its annotated target, as defined in Eq.~\eqref{eq:recall_routing_accuracy}.
\par
\end{samepage}

\begin{table}[htbp]
\centering
\small
\setlength{\tabcolsep}{2.0pt}
\begin{tabular}{@{}lccccc@{}}
\toprule
Type & Original & Shortened & Incomplete & Overlap & Mixed \\
\midrule
Text Sim. & 90.1\% & 85.0\% & 74.3\% & 78.5\% & 65.2\% \\
Recall Acc.\(\red{\uparrow}\) & 99.5\% & 93.4\% & 89.6\% & 92.0\% & 83.2\% \\
\bottomrule
\end{tabular}

\caption{Text similarity and recall-routing accuracy under prompt perturbations.}
\label{tab:ambiguous_prompt_routing}
\end{table}

Table~\ref{tab:ambiguous_prompt_routing} shows that recall accuracy ranges from 89.6\% to 93.4\% for individual perturbations. In the mixed setting, text similarity drops from 90.1\% to 65.2\%, and the router still retrieves the intended scene in 83.2\% of cases.

\section{Additional method details and ablations}
\label{app:ablation}

\subsection{Bidirectional rolling early anchors}

Early frames are generated within the model's training horizon and usually provide cleaner global references. Echo-Forcing therefore maintains a rolling anchor pool $\mathcal{P}_{\mathrm{roll}}$ with $|\mathcal{P}_{\mathrm{roll}}|=18$, from which $N_{\mathrm{anc}}=12$ active anchors are used in the cache. Each update inserts a block of $S=3$ anchor frames. Instead of using a fixed order, we traverse the pool in alternating forward and backward directions, which refreshes early references while avoiding discontinuities at cycle boundaries.

\begin{table*}[t]
\centering
\small
\setlength{\tabcolsep}{1.8pt}
\renewcommand{\arraystretch}{1.2}
\begin{tabular}{lc|cccccccc}
\toprule
Method
&
& \begin{tabular}{@{}c@{}}Aesthetic\\ Quality$\textcolor{red}{\uparrow}$\end{tabular}
& \begin{tabular}{@{}c@{}}Background\\ Consistency$\textcolor{red}{\uparrow}$\end{tabular}
& \begin{tabular}{@{}c@{}}Imaging\\ Quality$\textcolor{red}{\uparrow}$\end{tabular}
& \begin{tabular}{@{}c@{}}Subject\\ Consistency$\textcolor{red}{\uparrow}$\end{tabular}
& \begin{tabular}{@{}c@{}}Motion\\ Smoothness$\textcolor{red}{\uparrow}$\end{tabular}
& \begin{tabular}{@{}c@{}}Temporal\\ Flickering$\textcolor{red}{\uparrow}$\end{tabular}
& \begin{tabular}{@{}c@{}}Dynamic\\ Degree$\textcolor{red}{\uparrow}$\end{tabular}
& \begin{tabular}{@{}c@{}}Overall\\ Avg.$\textcolor{red}{\uparrow}$\end{tabular}
\\
\midrule

Static
& &
60.32 & \underline{96.32} & 71.90 & 97.22 & \textbf{98.91} & \underline{97.76} & 27.08
& 78.50
\\

Positive
& &
\underline{61.21} & 96.02 & 71.86 & \underline{97.61} & 98.35 & 96.98 & \underline{42.50}
& 80.65
\\

Reverse
& &
60.81 & 96.05 & \underline{71.94} & \textbf{97.83} & 98.77 & 97.63 & 42.08
& \underline{80.73}
\\

\rowcolor{SOTARow}
Bidirectional (Ours)
& &
\textbf{61.69} & \textbf{97.17} & \textbf{72.09} & 97.17
& \underline{98.79} & \textbf{98.28} & \textbf{47.59}
& \textbf{81.83}
\\
\bottomrule
\end{tabular}
\caption{Ablation of rolling-anchor update mode on the 60-second VBench-Long protocol.  We compare static anchors, one-directional rolling, and bidirectional rolling under the same setting as the main long-video results.}
\label{tab:rolling_mode}
\end{table*}

\paragraph{Effect of rolling strategy.}
Table~\ref{tab:rolling_mode} shows that static anchors provide stable references but strongly suppress motion, yielding a low dynamic degree of 27.08. Forward and reverse rolling improve dynamic degree to 42.50 and 42.08, respectively, but introduce weaker temporal stability around rolling boundaries. Bidirectional rolling achieves the best dynamic degree of \textbf{47.59} and the best temporal flickering score of \textbf{98.28}, while also improving background consistency to \textbf{97.17}. This confirms that alternating anchor traversal better preserves motion without sacrificing long-range stability.

\begin{table*}[t]
\centering
\small
\setlength{\tabcolsep}{2.4pt}
\renewcommand{\arraystretch}{1.2}

\begin{tabular}{lc|cccccccc}
\toprule
Sink & Compress
& \begin{tabular}{@{}c@{}}Aesthetic\\ Quality$\textcolor{red}{\uparrow}$\end{tabular}
& \begin{tabular}{@{}c@{}}Background\\ Consistency$\textcolor{red}{\uparrow}$\end{tabular}
& \begin{tabular}{@{}c@{}}Imaging\\ Quality$\textcolor{red}{\uparrow}$\end{tabular}
& \begin{tabular}{@{}c@{}}Subject\\ Consistency$\textcolor{red}{\uparrow}$\end{tabular}
& \begin{tabular}{@{}c@{}}Motion\\ Smoothness$\textcolor{red}{\uparrow}$\end{tabular}
& \begin{tabular}{@{}c@{}}Temporal\\ Flickering$\textcolor{red}{\uparrow}$\end{tabular}
& \begin{tabular}{@{}c@{}}Dynamic\\ Degree$\textcolor{red}{\uparrow}$\end{tabular}
& \begin{tabular}{@{}c@{}}Overall\\ Avg.$\textcolor{red}{\uparrow}$\end{tabular}
\\
\midrule

6  & 9
& 61.29 & 96.01 & \textbf{72.42} & 97.08 & 98.72 & 97.71 & 39.27
& 80.36
\\

9  & 6
& 61.52 & 96.03 & 71.47 & 97.16 & \underline{98.74} & \underline{97.86} & 37.39
& 80.02
\\

15 & 0
& \underline{61.65} & \underline{96.25} & 72.07 & \textbf{97.68} & 98.71 & 97.85 & \underline{41.04}
& \underline{80.75}
\\
\rowcolor{SOTARow}
12 & 3
& \textbf{61.69} & \textbf{97.17} & \underline{72.09} & \underline{97.17} & \textbf{98.79} & \textbf{98.28} & \textbf{47.59}
& \textbf{81.83}
\\

\bottomrule
\end{tabular}
\label{tab:sink_compress}
\caption{Memory-budget allocation on the 60-second VBench-Long protocol.  We vary the split between rolling anchors and compressed history under the fixed cache budget used in the main long-video evaluation.}
\label{tab:ablation_compress_size}
\end{table*}

\paragraph{Effect of memory budget allocation.}
Table~\ref{tab:ablation_compress_size} studies how the cache budget should be allocated between sink anchors and compressed history. Using more compressed history, such as $6$ anchors and $9$ compressed frames, improves imaging quality but weakens temporal stability and dynamics. Using only anchors, such as $15$ anchors and $0$ compressed frames, gives strong subject consistency but reduces dynamic degree to 41.04, indicating over-reliance on static early references. Our default allocation, $12$ anchors and $3$ compressed frames, achieves the best background consistency, motion smoothness, temporal flickering, and dynamic degree. This suggests that a small amount of compressed history is necessary to complement stable anchors with evolving long-range context.

\subsection{Drift-gated phase compression}
\label{sec:compression}
Drift-Gated Phase Compression selects informative long-range tokens using a stable calibrated query center while adapting to recent query drift. Directly relying on recent queries is sensitive to local noise and may greedily select tokens that are only useful for the current block. In contrast, the calibrated pre-RoPE query center provides a stable phase reference, and the drift gate controls how much amplitude compensation should be injected when the current query distribution deviates from the calibration stage.

Compressed tokens are assigned the timestamp of the current block rather than retaining their original temporal indices. This avoids timestamp aliasing within the compressed region and keeps attention computation phase-consistent. We fix the candidate compression region to $N=18$ frames. The calibrated query statistic $\bar{q}$ is computed per attention head and spatial position, with size $[H,d_{\mathrm{head}}]$, so the scoring overhead is lightweight and scales linearly with the candidate region.

\begin{table}[t]
\centering
\small
\setlength{\tabcolsep}{1.2pt}
\renewcommand{\arraystretch}{1.1}
\begin{tabular}{l|cccc}
\toprule
Method
& \begin{tabular}{@{}c@{}}Aesthetic\\ Quality$\textcolor{red}{\uparrow}$\end{tabular}
& \begin{tabular}{@{}c@{}}Background\\ Consist.$\textcolor{red}{\uparrow}$\end{tabular}
& \begin{tabular}{@{}c@{}}Dynamic\\ Degree$\textcolor{red}{\uparrow}$\end{tabular}
& \begin{tabular}{@{}c@{}}Overall\\ Avg.$\textcolor{red}{\uparrow}$\end{tabular}
\\
\midrule

$\lambda = 1$
& 60.95
& 95.97
& \underline{36.35}
& \underline{64.42}
\\

$\lambda = 3$
& \underline{61.64}
& \underline{96.07}
& 34.79
& 64.17
\\
\rowcolor{SOTARow}
Ours ($\lambda = 2$)
& \textbf{61.69}
& \textbf{97.17}
& \textbf{47.59}
& \textbf{68.82}
\\

\bottomrule
\end{tabular}
\caption{Sensitivity to \(\lambda\) on the 60-second VBench-Long protocol.  The displayed metrics use the same experiment as the main long-video evaluation; Selected Avg. is the arithmetic mean of the displayed metrics.}
\label{tab:ablation_gate_value}
\end{table}

\paragraph{Effect of drift-gate sensitivity coefficient.}
Table~\ref{tab:ablation_gate_value} shows that the drift-gate sensitivity coefficient controls the trade-off between stability and adaptability. A small \(\lambda\) under-reacts to query drift, leading to weaker background consistency and dynamic degree. A large \(\lambda\) suppresses amplitude compensation too aggressively, also reducing dynamic degree. The default setting, \(\lambda=2\), achieves the best background consistency of \textbf{97.17}, motion smoothness of \textbf{98.79}, temporal flickering of \textbf{98.28}, and dynamic degree of \textbf{47.59}. This indicates that moderate drift-gate sensitivity is important for retaining useful historical dynamics while filtering mismatched memories.

\subsection{Scene Recall Frames}
\label{sec:recall}
Scene Recall Frames provide compact scene-level memories for long-range recall. Storing all frames from a scene is costly and may inject redundant or noisy old-scene information. Selecting a single frame is also insufficient, because different frames may preserve complementary details such as subject texture, occluded regions, or background layout. We therefore sample \(M=5\) candidate frames from the stable part of each scene and fuse their KV states independently at each spatial position. Each scene is stored as one recall frame, and the scene-memory pool has a capacity of 64 frames; excess entries are evicted in FIFO order. The pool capacity can be increased when a larger recall context is required.

\begin{table*}[t]
\centering
\captionsetup[subtable]{justification=centering}
\begin{subtable}[t]{0.50\linewidth}
\vspace{0pt}
\centering
\small
\setlength{\tabcolsep}{1.8pt}
\renewcommand{\arraystretch}{1.2}
\begin{tabular}{l|ccccc}
\toprule
Method
& \begin{tabular}{@{}c@{}}Text\\ Align$\textcolor{red}{\uparrow}$\end{tabular}
& \begin{tabular}{@{}c@{}}Subject\\ Consist.$\textcolor{red}{\uparrow}$\end{tabular}
& \begin{tabular}{@{}c@{}}Aesthetic\\ Quality$\textcolor{red}{\uparrow}$\end{tabular}
& \begin{tabular}{@{}c@{}}Imaging\\ Quality$\textcolor{red}{\uparrow}$\end{tabular}
& \begin{tabular}{@{}c@{}}Overall\\ Avg.$\textcolor{red}{\uparrow}$\end{tabular}
\\
\midrule

No memory
& 31.14 & 75.12 & 53.65 & \underline{68.43}
& 57.09
\\

First Frame
& 32.21 & 80.22 & \underline{54.61} & 68.21
& 58.81
\\

Key Frame
& \underline{32.24} & \underline{80.35} & 54.52 & 68.30
& \underline{58.85}
\\

\rowcolor{SOTARow}
Recall Frame (ours)
& \textbf{32.58} & \textbf{83.48} & \textbf{54.98} & \textbf{69.13}
& \textbf{60.04}
\\

\bottomrule
\end{tabular}
\caption{Recall source under the fine-tuned Scene Recall protocol.}
\label{tab:ablation_recall_source}
\end{subtable}%
\hfill%
\begin{subtable}[t]{0.49\linewidth}
\vspace{0pt}
\centering
\small
\setlength{\tabcolsep}{3.0pt}
\renewcommand{\arraystretch}{1.03}
\begin{tabular}{l|cccc}
\toprule
Method
& \begin{tabular}{@{}c@{}}Text\\ Align$\textcolor{red}{\uparrow}$\end{tabular}
& \begin{tabular}{@{}c@{}}Subject\\ Consist.$\textcolor{red}{\uparrow}$\end{tabular}
& \begin{tabular}{@{}c@{}}Background\\ Consist.$\textcolor{red}{\uparrow}$\end{tabular}
& \begin{tabular}{@{}c@{}}Overall\\ Avg.$\textcolor{red}{\uparrow}$\end{tabular}
\\
\midrule

No decay
& 25.74 & 94.68 & 92.45
& 70.96
\\

decay = 0.90
& 26.08 & \underline{94.92} & 92.62
& 71.21
\\

decay = 0.75
& 26.44 & 94.60 & 92.93
& 71.32
\\

decay = 0.50
& \underline{27.34}
& 94.77
& \underline{93.45}
& \underline{71.85}
\\

\rowcolor{SOTARow}
Ours
& \textbf{29.77}
& \textbf{95.32}
& \textbf{93.74}
& \textbf{72.94}
\\

\bottomrule
\end{tabular}
\caption{Memory decay under the fine-tuned smooth-transition protocol.}
\label{tab:ablation_decay_speed}
\end{subtable}
\caption{Ablation of interactive scene-memory modules.  The recall-source results use the same Scene Recall experiment as the fine-tuned main result, and the memory-decay results use the same smooth-transition experiment.}
\label{tab:ablation_recall_decay}
\end{table*}

\paragraph{Effect of recall source.}
Table~\ref{tab:ablation_recall_source} compares different memory sources for scene recall. Without memory, the model lacks access to earlier scene priors and obtains only 75.12 subject consistency. Selecting the first frame or a single crucial frame improves subject consistency slightly, but still cannot capture multi-frame complementary information. Our Scene Recall Frames increases subject consistency to \textbf{83.48} and text alignment to \textbf{32.58}, clearly outperforming single-frame alternatives. Its video quality remains comparable to the best single-frame baseline, showing that the gain in recall fidelity does not come at the cost of overall visual quality.

\paragraph{Single-frame baselines and metrics.}
For the \emph{First Frame} baseline, we directly store the KV states of the first stable block in each historical scene. For the \emph{Key Frame} baseline, we select one candidate block whose spatial keys are most compatible with the scene-level calibrated query representation:
\begin{equation}
j_s^{\star}
=
\arg\max_{j\in\{1,\ldots,M\}}
\frac{1}{|\mathcal{U}|}
\sum_{u\in\mathcal{U}}
\operatorname{sim}
\left(
\bar{\mathbf{q}}_{s,u},
\mathbf{k}^{\mathrm{raw}}_{s,j,u}
\right),
\end{equation}
where $\mathcal{U}$ denotes the set of spatial positions. The selected block's spatial KV tokens are directly stored as the scene memory and injected at recall time. Both single-frame baselines use the same routing decision and cache budget as our method; their only difference is that they retain one unaggregated block rather than fusing multiple candidate blocks.

For this ablation, Text Align. measures segment-wise alignment between each generated segment and its matched prompt, whereas Subject Consist. measures identity preservation in the recalled scene. Aesthetic Quality and Imaging Quality assess the visual quality of the recalled output. Overall Avg. is the arithmetic mean of these four metrics. This protocol separates scene-memory fidelity from prompt following and visual quality.

\subsection{Difference-aware Memory Decay}
\label{sec:decay}

After a scene transition, the KV cache may still contain tokens from the previous scene. 
These old-scene tokens can be useful when they share compatible subject or background information with the new scene, but they may also introduce residual semantics when the new prompt requires a different background, action, or layout. 
Therefore, instead of directly flushing all old memories, Echo-Forcing applies a token-wise soft forgetting mechanism that adaptively suppresses old tokens according to their discrepancy with the new-scene reference.

Let $(\mathbf{k}_i^{\mathrm{old}}, \mathbf{v}_i^{\mathrm{old}})$ denote the $i$-th old-memory token preserved after the transition. 
After entering the new scene, we first generate the initial clean block and use it as the new-scene reference. 
Let $\mathbf{k}_i^{\mathrm{new}}$ denote the key token at the corresponding or nearest spatial position in this reference block. 
We estimate the old--new discrepancy by cosine distance:
\begin{equation}
d_i
=
1-
\cos\left(\mathbf{k}_i^{\mathrm{old}}, \mathbf{k}_i^{\mathrm{new}}\right)
=
1-
\frac{
\left(\mathbf{k}_i^{\mathrm{old}}\right)^\top \mathbf{k}_i^{\mathrm{new}}
}{
\left\|\mathbf{k}_i^{\mathrm{old}}\right\|_2
\left\|\mathbf{k}_i^{\mathrm{new}}\right\|_2
}.
\end{equation}
The discrepancy scores are normalized within the old-memory set:
\begin{equation}
\hat{d}_i
=
\frac{
d_i-\min_j d_j
}{
\max_j d_j-\min_j d_j+\epsilon
},
\end{equation}
where $\epsilon$ is a small constant for numerical stability. 
We then map the normalized discrepancy to a token-wise decay strength:
\begin{equation}
\mu_i
=
\mu_{\min}
+
(\mu_{\max}-\mu_{\min})\hat{d}_i .
\end{equation}
In this way, tokens that are more inconsistent with the new scene receive larger decay strengths, while tokens that remain compatible with the new scene decay more slowly.

At the $r$-th generation step after the transition, we define the memory weight of token $i$ as
\begin{equation}
w_i^{(r)}
=
\exp(-r\mu_i),
\qquad
0 < w_i^{(r)} \le 1 .
\end{equation}
The old key and value are then scaled as
\begin{equation}
\tilde{\mathbf{k}}_i^{(r)}
=
w_i^{(r)} \mathbf{k}_i^{\mathrm{old}},
\qquad
\tilde{\mathbf{v}}_i^{(r)}
=
w_i^{(r)} \mathbf{v}_i^{\mathrm{old}} .
\end{equation}

For a query $\mathbf{q}$, the attention logit between $\mathbf{q}$ and the decayed old key becomes
\begin{equation}
\tilde{e}_i^{(r)}
=
\frac{\mathbf{q}^\top \tilde{\mathbf{k}}_i^{(r)}}{\sqrt{d}}
=
w_i^{(r)}
\frac{\mathbf{q}^\top \mathbf{k}_i^{\mathrm{old}}}{\sqrt{d}}
=
w_i^{(r)} e_i^{\mathrm{old}},
\end{equation}
where $e_i^{\mathrm{old}} = \mathbf{q}^\top \mathbf{k}_i^{\mathrm{old}}/\sqrt{d}$ is the original attention logit.
For a fixed query and old key, decreasing $w_i^{(r)}$ scales the logit toward zero, with absolute value $|\tilde{e}_i^{(r)}| = w_i^{(r)} |e_i^{\mathrm{old}}|$.
The corresponding softmax attention weight is
\begin{equation}
\tilde{\mathbf{a}}_i^{(r)}
=
\frac{
\exp\left(\tilde{e}_i^{(r)}\right)
}{
\sum_{\ell \in \mathcal{M}_{\mathrm{old}}}
\exp\left(\tilde{e}_\ell^{(r)}\right)
+
\sum_{j \in \mathcal{M}_{\mathrm{new}}}
\exp\left(e_j^{\mathrm{new}}\right)
},
\end{equation}
where $\mathcal{M}_{\mathrm{old}}$ and $\mathcal{M}_{\mathrm{new}}$ denote the old-memory tokens and the newly generated tokens, respectively.
The direction of change in the softmax attention weight depends on the original logit and on the other logits in the denominator.

The attention output uses the scaled values
\begin{equation}
\tilde{o}^{(r)}
=
\sum_{i \in \mathcal{M}_{\mathrm{old}}}
\tilde{\mathbf{a}}_i^{(r)} \tilde{\mathbf{v}}_i^{(r)}
+
\sum_{j \in \mathcal{M}_{\mathrm{new}}}
\tilde{\mathbf{a}}_j^{(r)} \mathbf{v}_j^{\mathrm{new}} .
\end{equation}
For an old token $i$, its contribution is
\begin{equation}
\tilde{\mathbf{a}}_i^{(r)} \tilde{\mathbf{v}}_i^{(r)}
=
\tilde{\mathbf{a}}_i^{(r)} w_i^{(r)} \mathbf{v}_i^{\mathrm{old}} .
\end{equation}
At a fixed attention weight, value scaling reduces the magnitude of the old token's contribution by the factor $w_i^{(r)}$.

Overall, Difference-aware Memory Decay provides a soft alternative to hard cache flushing. 
It preserves compatible old memories during the early stage of a transition, which helps maintain subject or background continuity when useful, while rapidly suppressing inconsistent regions that may contaminate the new scene. 
This token-wise decay enables Echo-Forcing to handle both smooth transitions and hard cuts within a unified memory-update process.

\paragraph{Effect of decay strategy.}
Table~\ref{tab:ablation_decay_speed} shows that fixed decay improves over no decay, but a single global rate cannot adapt to spatially different old-new conflicts. Stronger fixed decay improves text alignment from 25.74 to \textbf{27.34}, but remains limited because it also suppresses useful consistent regions. Our difference-aware decay achieves \textbf{29.77} text alignment,\textbf{ 95.32} subject consistency, and \textbf{93.74} background consistency. Compared with the best fixed decay, it improves text alignment by 2.43 points while also improving both subject and background consistency. This confirms that spatially adaptive forgetting better removes conflicting old-scene semantics while preserving reusable priors.

\subsection{Relative RoPE extrapolation}
\label{sec:relative-rope}

For non-fine-tuned backbones such as Self-Forcing~\citep{huang2025self}, the training window is limited to $L=21$ frames.
Directly applying RoPE with continuously increasing absolute temporal indices would expose the model to positions far beyond its training range, leading to severe length-extrapolation artifacts.
We therefore adopt relative-time RoPE, which re-encodes cached KV states by mapping the active cache into a bounded temporal interval.

Specifically, let $t$ denote the absolute frame index during autoregressive rollout, and let $\mathcal{T}_b=\{t_1,t_2,\ldots,t_n\}$ denote the frame indices stored in the active cache at generation block $b$, where $n\leq L$.
Instead of using the absolute index $t_i$, we assign each cached frame a relative RoPE index
\begin{equation}
\rho_b(t_i)
=
i-1,
\qquad
i=1,\ldots,n,
\qquad
\rho_b(t_i)\in[0,L-1].
\end{equation}
Equivalently, for a cache ordered from old to recent, the oldest cached frame is mapped to $0$ and the newest cached frame is mapped to $n-1\leq L-1$.
The newly generated frame is then assigned the next local index within the same bounded window.
Thus, all RoPE temporal coordinates remain within the training range of $21$ frames, even when the absolute rollout length grows to hundreds or thousands of frames.

This relative re-indexing preserves local temporal distances inside the active cache while avoiding unbounded absolute RoPE positions.
It allows Echo-Forcing to perform long autoregressive rollout without changing the pretrained model parameters.
For fine-tuned long-video backbones such as LongLive~\citep{yang2025longlive}, we follow their native temporal encoding configuration to preserve their learned long-range modeling ability.

\subsection{Computation and memory overhead}

Echo-Forcing introduces only lightweight computation on top of autoregressive video generation.
The dominant cost still comes from the local denoising window of the base causal generator.
By default, we use a local cache of $L=21$ frames, which provides sufficient temporal context for stable long-video generation.
Therefore, compared with methods using smaller local windows, Echo-Forcing may introduce modest latency overhead, but its cost remains bounded and does not grow with the total video length.

Let $N_{\mathrm{cand}}$ denote the number of candidate tokens considered for compression, $M$ the number of candidate frames used for scene recall, and $B$ the fixed active memory budget.
The additional overhead of Echo-Forcing is
\begin{equation}
\mathcal{O}_{\mathrm{extra}}
=
\mathcal{O}(N_{\mathrm{cand}} + M + B),
\end{equation}
where the three terms correspond to drift-gated phase compression, scene recall frame construction, and transition-time memory decay, respectively.
All these quantities are bounded by the fixed cache budget rather than the total generated sequence length.
Moreover, online query calibration only maintains running statistics, and rolling-anchor updates are constant-level operations.

\paragraph{The cache distributions of representative methods.}
For compact notation, let $S$, $R$, $C$, and $A$ denote sink, recent, compressed, and anchor frames, respectively. At every generation step, three current-generation frames are retained by all methods. We include these shared frames in the recent component $R$, rather than listing them separately; thus, $R$ denotes the complete recent context, including the three current-generation frames and any additional recent cached frames.

Under the 21-frame cache setting, Self-Forcing~\citep{huang2025self} and $\infty$-RoPE~\citep{yesiltepe2025infinity} use pure recent caches, i.e., $\mathcal{M}_{\mathrm{SF}}=R_{21}$ and $\mathcal{M}_{\infty\text{-RoPE}}=R_{21}$. Rolling-Sink~\citep{li2026rolling} uses $\mathcal{M}_{\mathrm{RollingSink}}=S_{15}\oplus R_{6}$, Deep-Forcing~\citep{yi2025deep} uses $\mathcal{M}_{\mathrm{DeepForcing}}=S_{12}\oplus C_{9}$, and Echo-Forcing adopts $\mathcal{M}_{\mathrm{Echo}}=A_{12}\oplus C_{3}\oplus R_{6}$. Echo-Forcing therefore allocates the fixed budget to stable anchors, compressed history, and recent continuity.

LongLive~\citep{yang2025longlive} follows its native active-cache configuration, $\mathcal{M}_{\mathrm{LongLive}}=S_{3}\oplus R_{9}$, which contains 12 frames. We report this configuration separately rather than treating it as a 21-frame allocation.

Compared with Deep-Forcing, which allocates a larger compressed-history portion $C_{9}$, Echo-Forcing uses a smaller compressed cache $C_{3}$ and lightweight phase-based selection, leading to faster inference among history-compression methods \textbf{(16.01 vs.\ 15.65)}. Compared with the other 21-frame cache configurations, the extra operations of Echo-Forcing are marginal, while the structured allocation into anchors, compressed history, and recent frames brings substantial gains in long-horizon stability and interactive scene control.

LongLive achieves an inference speed of \textbf{16.41} FPS in our implementation under its native 12-frame active cache. The higher throughput reported in the LongLive paper (20.70 FPS) also uses this smaller active-cache configuration, $\mathcal{M}_{\mathrm{LongLive}}=S_{3}\oplus R_{9}$.

Since all memory operations are performed under a bounded cache, Echo-Forcing avoids the unbounded memory growth of full-history caching while preserving compact long-range scene memories beyond a simple sliding window.

\section{Additional visualizations}
\label{app:visualizations}

\begin{figure*}[!t]
    \centering
    \includegraphics[width=0.70\textwidth]{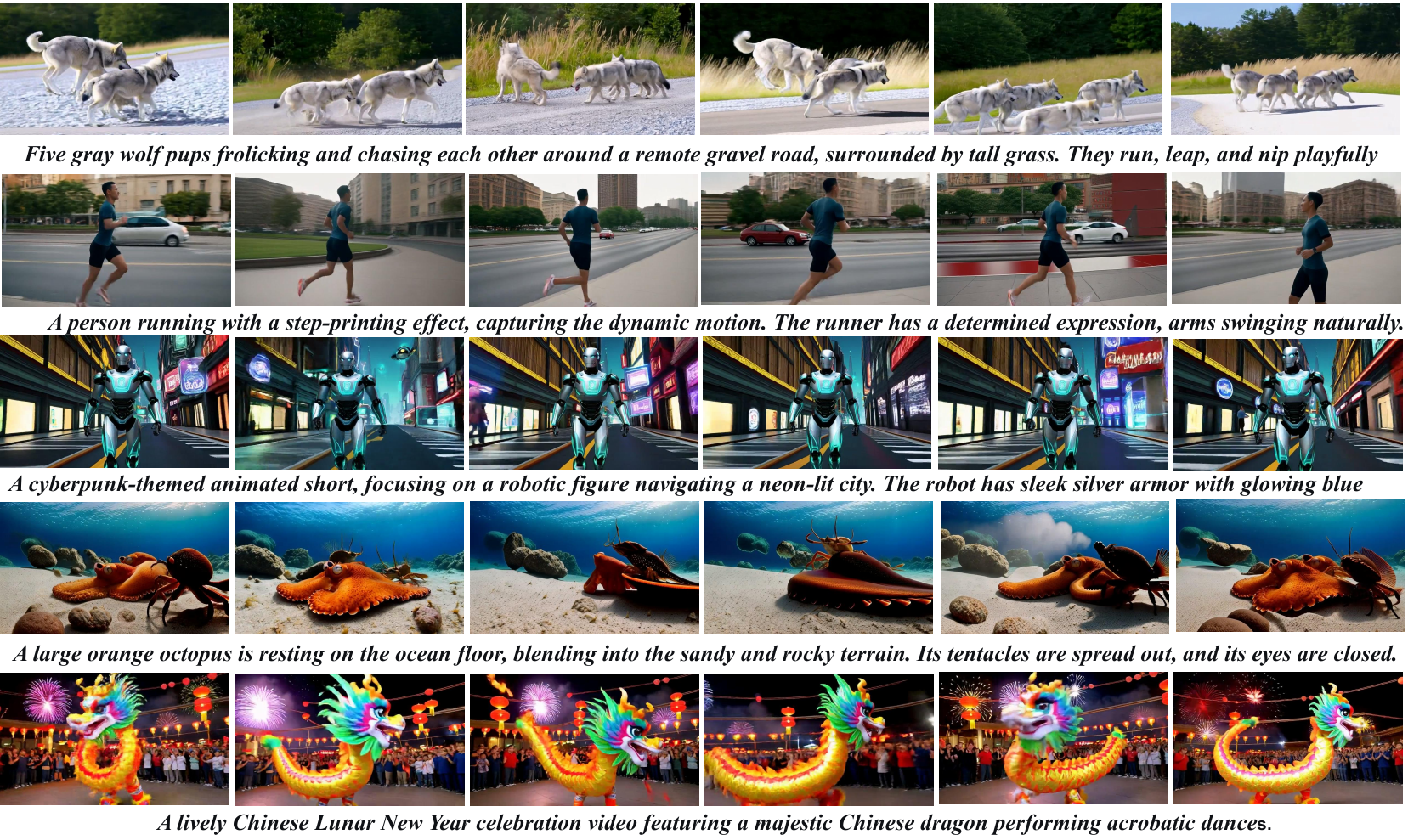}
    \caption{
    Additional visualization of long-video generation.
    Echo-Forcing maintains subject identity, background structure, and visual fidelity over extended autoregressive rollouts.
    }
    \label{fig:app_add_long}

    \par\medskip
    \includegraphics[width=0.70\textwidth]{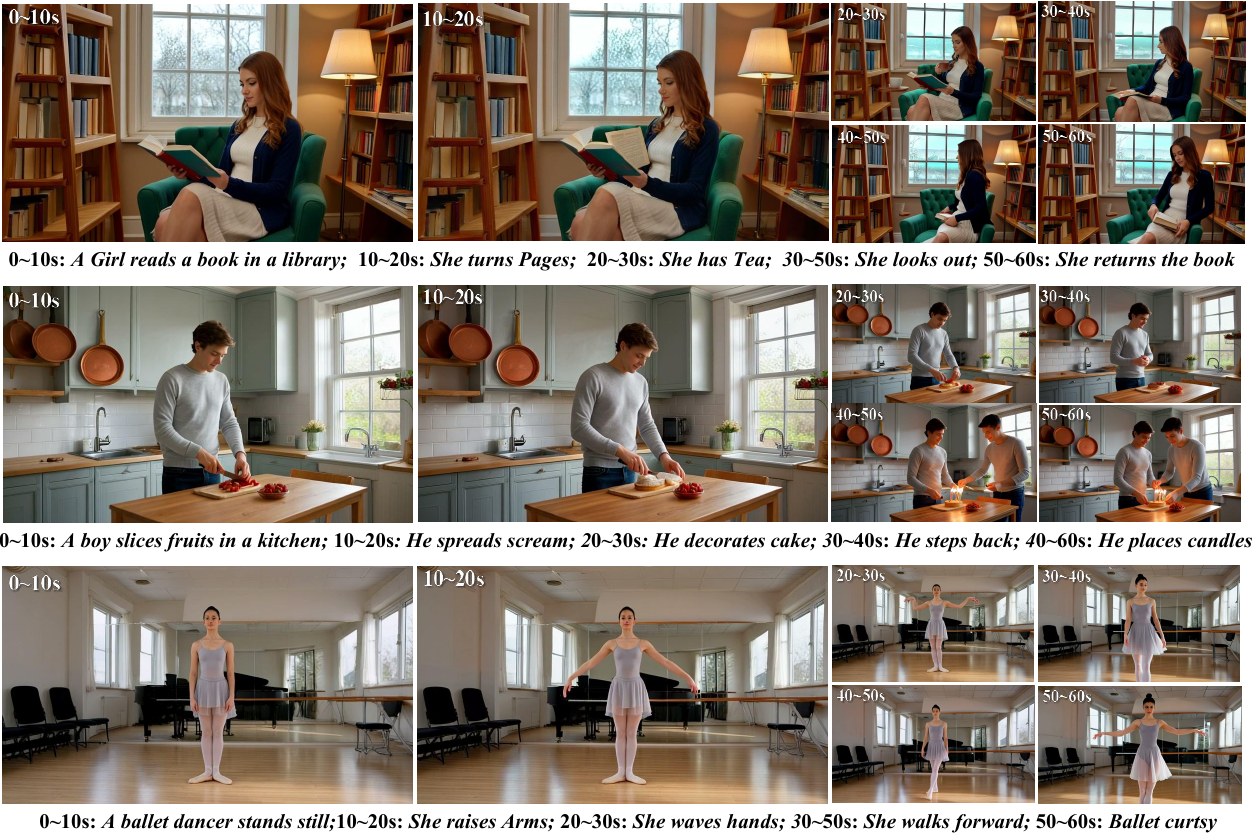}
    \caption{
    Additional visualization of smooth transitions.
    We show more examples of gradual prompt evolution under continuous scene dynamics. Echo-Forcing preserves compatible subject and scene priors across adjacent segments, producing smoother motion changes and more coherent visual transitions.
    }
    \label{fig:app_add_smooth}
\end{figure*}

\begin{figure*}[!tp]
    \centering
    \includegraphics[width=1.0\textwidth]{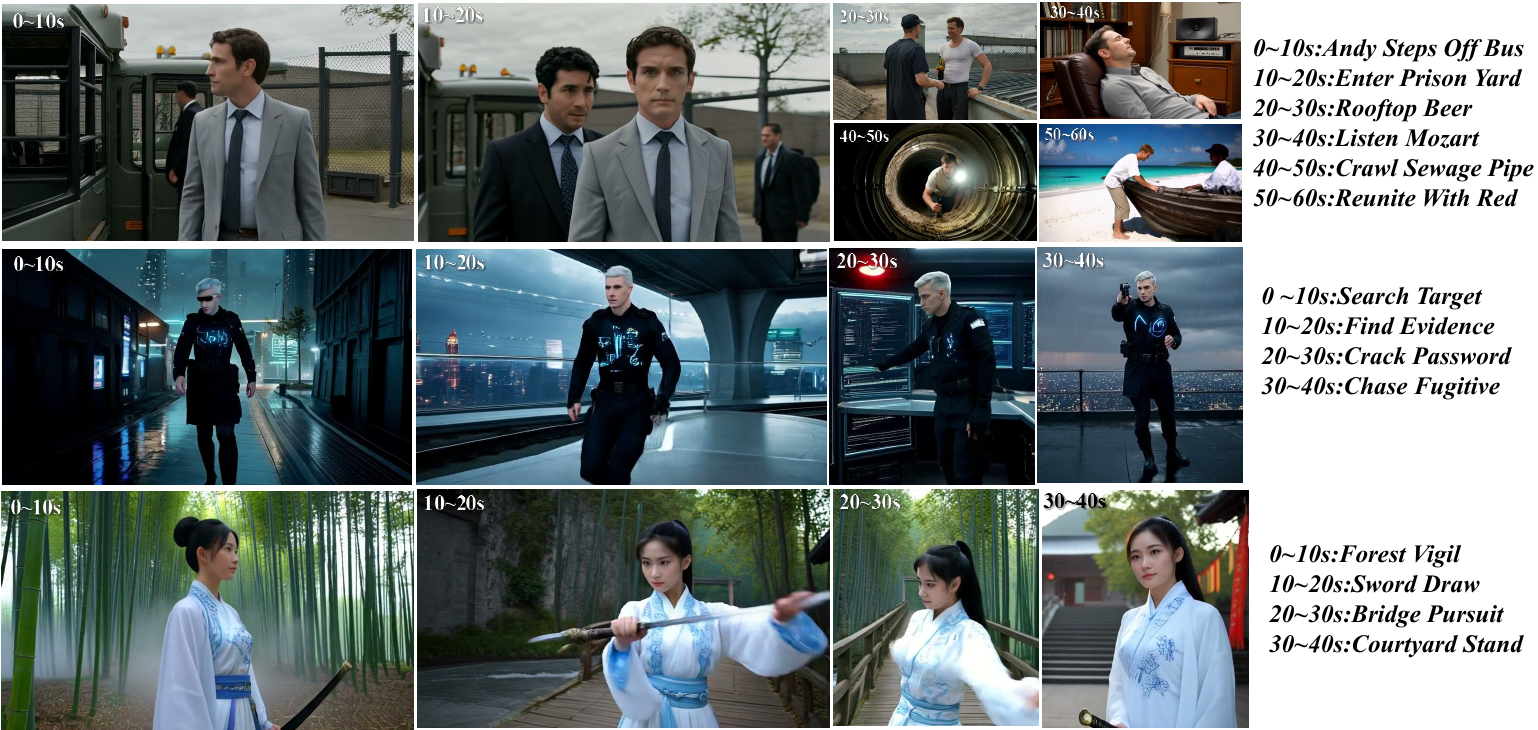}
    \caption{
    Additional visualization of hard cuts.
    We show more examples under abrupt semantic changes, where the subject is preserved while the background, action, or scene layout changes substantially. Echo-Forcing suppresses old-scene residuals and adapts more cleanly to the new prompt after each cut.
    }
    \label{fig:app_add_hard}

    \par\medskip
    \includegraphics[width=0.96\textwidth]{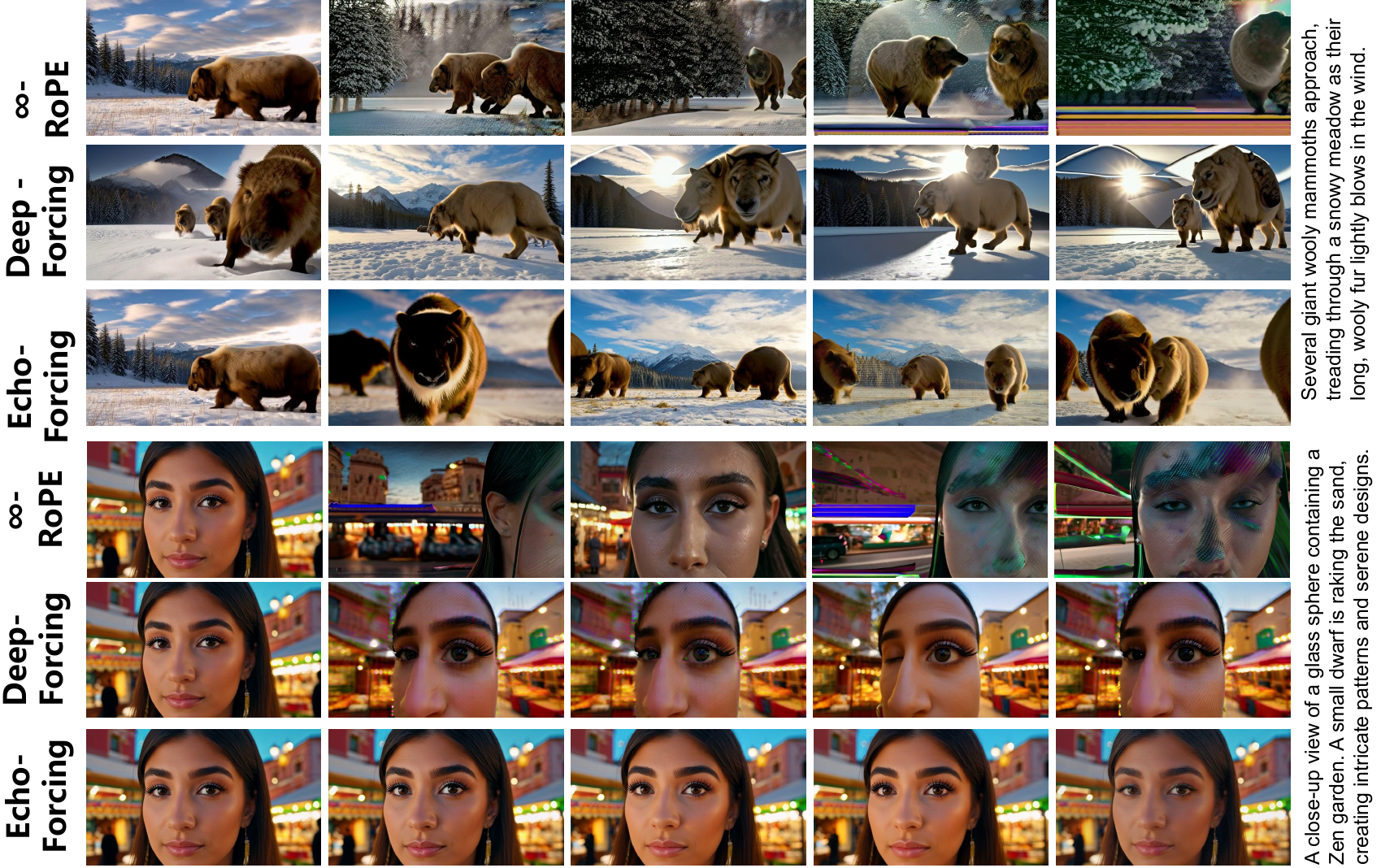}
    \caption{
    Qualitative comparison on 2-minute long-video generation.
    Echo-Forcing better preserves subject appearance, background coherence, and visual fidelity during 2-minute autoregressive rollout compared with representative baselines.
    }
    \label{fig:app_add_compare_long}
\end{figure*}

\begin{figure*}[!tp]
    \centering
    \includegraphics[width=0.90\textwidth]{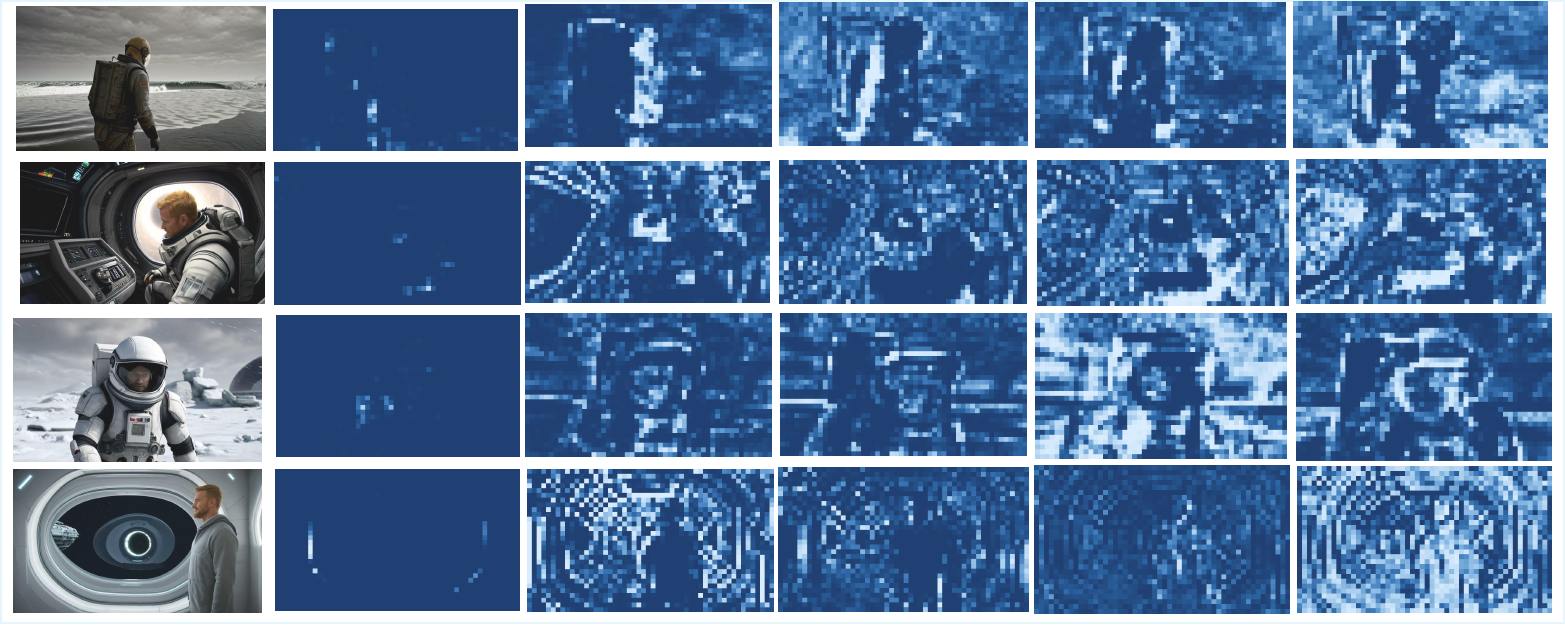}
    \caption{
    Visualization of Scene Recall Frames.
    Each row corresponds to one scene. The left image shows the scene reference, and the blue maps on the right visualize several recalled frames. Different recalled frames exhibit different temporal attention patterns, reflecting varying emphasis on scene information over time.
    }
    \label{fig:app_add_compress}

    \par\medskip
    \includegraphics[width=0.74\textwidth]{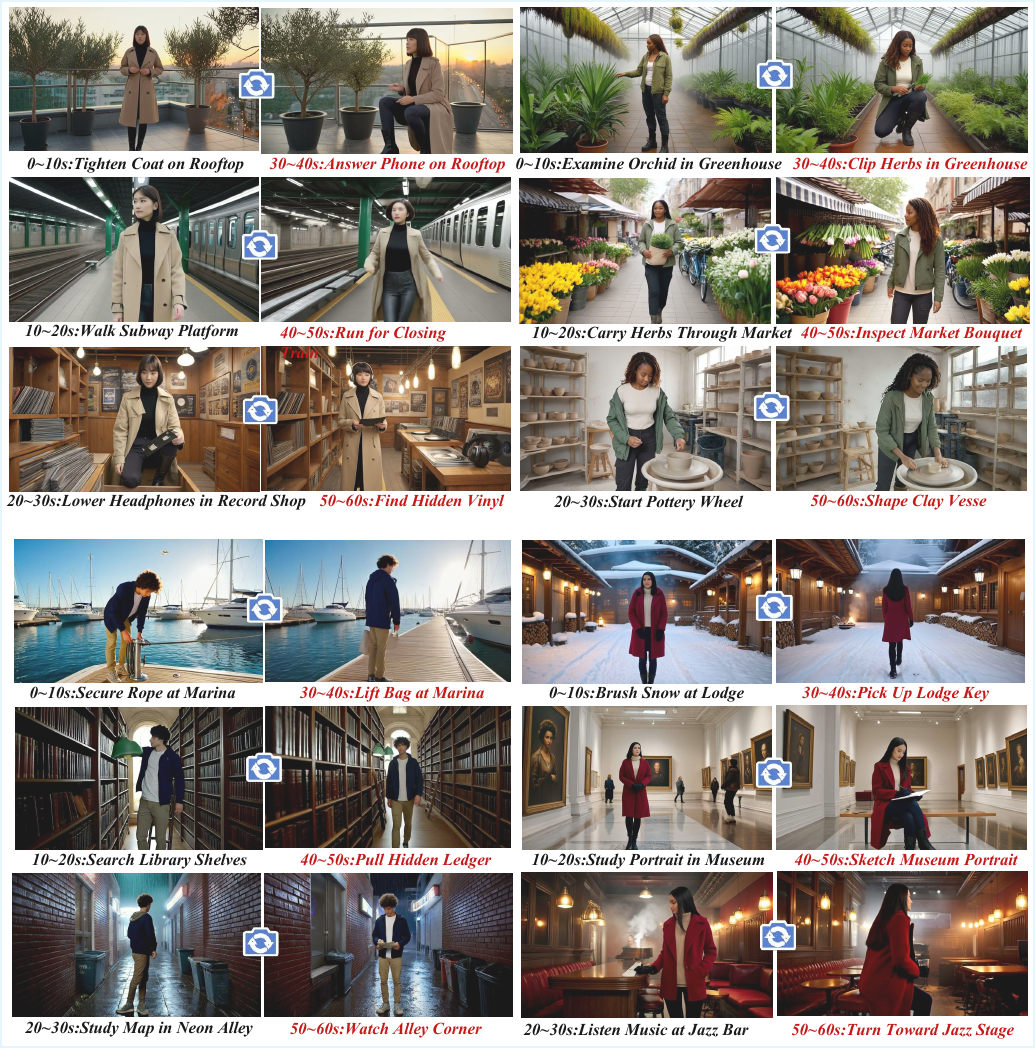}
    \caption{
    Additional scene-recall results.
    Echo-Forcing retrieves earlier scene memories and reduces semantic confusion across long-range shot intervals.
    }
    \label{fig:app_add_mem}
\end{figure*}

Figures~\ref{fig:app_add_long}--\ref{fig:app_add_mem} provide additional qualitative results across long-video generation, smooth transition, hard cut, scene recall, and historical-token compression.

\section{Limitations and Future Work}

\paragraph{Limitations.}
Echo-Forcing remains constrained by the capability of the underlying video generator and the finite memory budget available at inference time. Although the proposed memory lifecycle improves long-horizon consistency and interaction, it cannot fully eliminate errors accumulated during autoregressive generation. In addition, the heuristic scene-routing mechanism may be sensitive to ambiguous user instructions in open-ended interactive settings.

\paragraph{Future Work.}
Future work may explore learning-based memory policies that dynamically determine what to retain, compress, retrieve, or forget during long-horizon generation. Integrating richer visual grounding, such as object-level memory and reference-image conditioning, could further support multi-character narratives and finer-grained editing. It would also be valuable to develop more flexible interaction interfaces that combine natural-language instructions, optional control tags, and user feedback. Finally, evaluating interactive video generation on larger-scale open-ended scripts and real creative workflows may better characterize the practical utility of lifecycle-aware memory management.

\end{document}